%% file: main.tex
\documentclass[10pt,journal,compsoc]{IEEEtran}
\usepackage{cite}
\usepackage{graphicx}
\usepackage{balance}  
\usepackage{color}
\usepackage{amsfonts,amssymb,amsmath,bm}
\usepackage{makecell}
\usepackage{subfigure}
\usepackage{multirow}
\usepackage[ruled,vlined]{algorithm2e}
\usepackage{booktabs}
\usepackage{hyperref}
\usepackage{threeparttable}
\usepackage{bbding}
\usepackage{enumitem}
\usepackage{hyperref}
\usepackage[hyphenbreaks]{breakurl}

\newtheorem{definition}{Definition}

\newcommand{\rv}[1]{\textcolor{blue}}

\begin{document}
\title{Finding Materialized Models for Model Reuse}

\author{
Minjun~Zhao,
Lu~Chen,
Keyu~Yang,
Yuntao~Du,
Yunjun~Gao,~\IEEEmembership{Member,~IEEE}

\IEEEcompsocitemizethanks{
        \IEEEcompsocthanksitem M. Zhao, L. Chen (Corresponding Author), K. Yang and Y. Du are with the College of Computer Science, Zhejiang University, Hangzhou 310027, China, E-mail:\{minjunzhao, luchen, kyyang, ytdu\}@zju.edu.cn.
        \IEEEcompsocthanksitem Y. Gao is with Alibaba–Zhejiang University Joint Institute of Frontier Technologies, Hangzhou, China, the College of Computer Science, Zhejiang University, Hangzhou 310027, China, E-mail: gaoyj@zju.edu.cn.
}

}

\IEEEtitleabstractindextext{
\begin{abstract}
Materialized model query aims to find the most appropriate materialized model as the initial model for model reuse. It is the precondition of model reuse, and has recently attracted much attention. {Nonetheless, the existing methods suffer from the need to provide source data, limited range of applications, and inefficiency since they do not construct a suitable metric to measure the target-related knowledge of materialized models. To address this, we present \textsf{MMQ}, a source-data free, general, efficient, and effective materialized model query framework.} It uses a Gaussian mixture-based metric called separation degree to rank materialized models. For each materialized model, \textsf{MMQ} first vectorizes the samples in the target dataset into probability vectors by directly applying this model, then utilizes Gaussian distribution to fit for each class of probability vectors, and finally uses separation degree on the Gaussian distributions to measure the target-related knowledge of the materialized model. Moreover, we propose an improved \textsf{MMQ} (\textsf{I-MMQ}), which significantly reduces the query time while retaining the query performance of \textsf{MMQ}. Extensive experiments on a range of practical model reuse workloads demonstrate the effectiveness and efficiency of \textsf{MMQ}.
\end{abstract}

}

\maketitle

\IEEEdisplaynontitleabstractindextext

\IEEEpeerreviewmaketitle

\input{introduction}
\input{preliminaries}
\input{framework}

\input{techniques}
\input{experiment}
\input{relatedwork}
\input{conclusion}
\input{acknowledgments}


\balance

\bibliographystyle{abbrv}
\bibliography{refer}

\vspace*{-6ex}
\begin{IEEEbiography}[{\includegraphics[width=0.96in,height=1.1in,clip,keepaspectratio]
{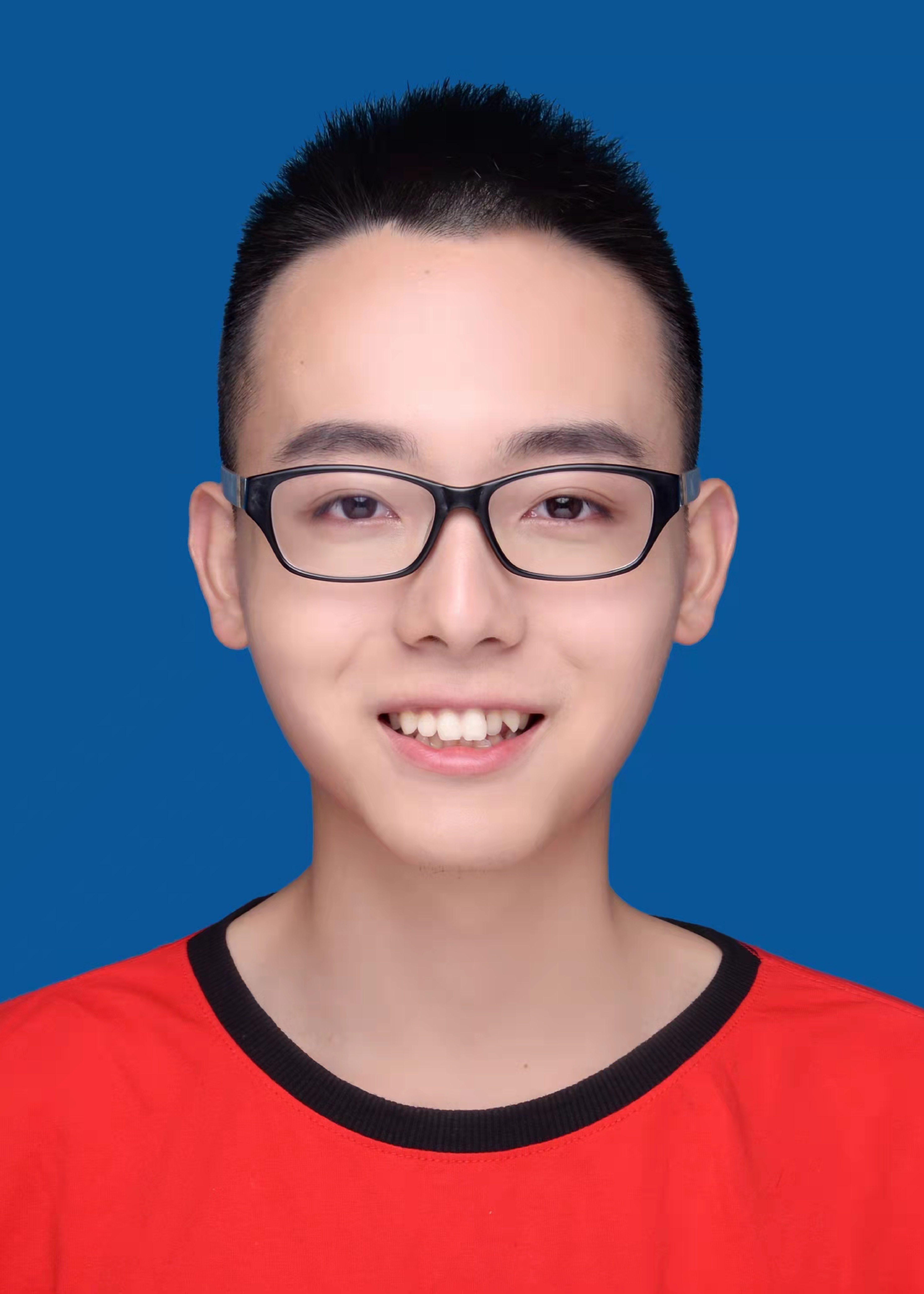}}]{Minjun Zhao}
received the BS degree in information security from Hainan University, China, in 2019. He is currently working toward the PhD degree in the College of Computer Science, Zhejiang University, China. His research interests include data management for machine learning and data mining.
\end{IEEEbiography}

\vspace*{-6ex}
\begin{IEEEbiography}[{\includegraphics[width=0.96in,height=1.1in,clip,keepaspectratio]
{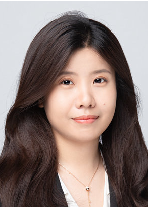}}]{Lu Chen}
received the PhD degree in computer science from Zhejiang University, China, in 2016. She was an assistant professor in Aalborg University for a 2-year period from 2017 to 2019, and she was an associated professor in Aalborg University for a 1-year period from 2019 to 2020. She is currently a ZJU Plan 100 Professor in the College of Computer Science, Zhejiang University, Hangzhou, China. Her research interests include indexing and querying metric spaces, graph databases, and graph mining.
\end{IEEEbiography}

\vspace*{-6ex}
\begin{IEEEbiography}[{\includegraphics[width=0.96in,height=1.1in,clip,keepaspectratio]
{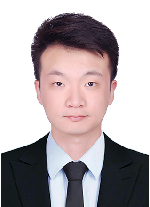}}]{Keyu Yang}
received the PhD degree in computer science from Zhejiang University, China, in 2021.  He is currently a Senior Algorithm Engineer with Huawei Technologies, China. His research interests include machine learning interaction with data management technology, and metric data management.
\end{IEEEbiography}

\vspace*{-6ex}
\begin{IEEEbiography}[{\includegraphics[width=0.96in,height=1.1in,clip,keepaspectratio]
{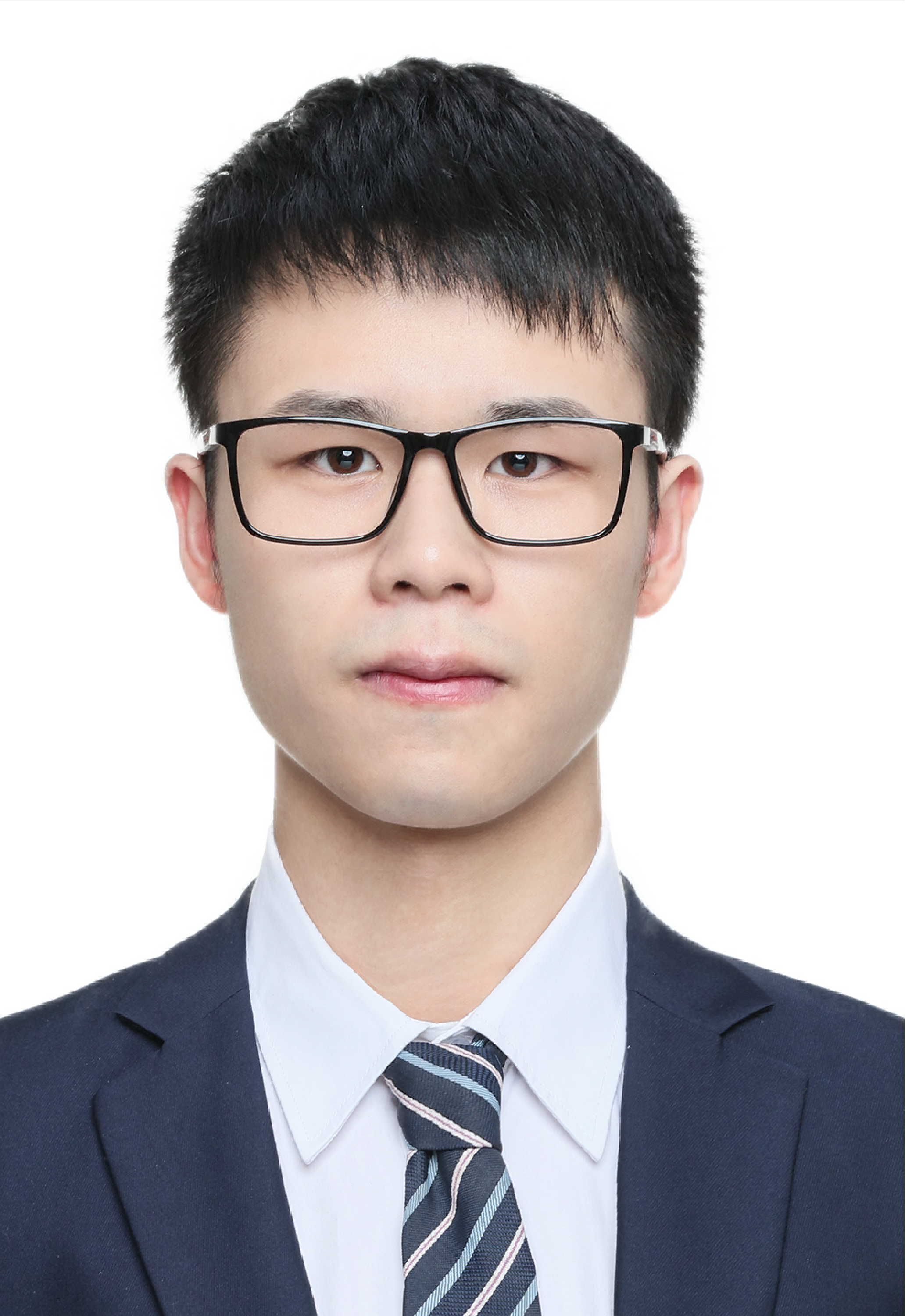}}]{Yuntao Du}
received the BS degree in data science from East China Normal University, China, in 2020. He is currently working toward the Master degree in the College of Computer Science, Zhejiang University, China. His research interests include recommender system and spatiotemporal data mining.
\end{IEEEbiography}

\vspace*{-6ex}
\begin{IEEEbiography}[{\includegraphics[width=0.96in,height=1.1in,clip,keepaspectratio]
{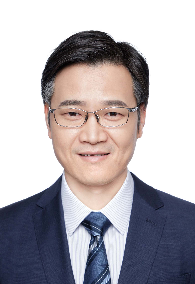}}]{Yunjun Gao}
received the PhD degree in computer science from Zhejiang University, China, in 2008. He is currently a professor in the College of Computer Science, Zhejiang University, China. His research interests include spatial and spatio-temporal databases, metric and incomplete/uncertain data management, graph databases, spatio-textual data processing, and database usability. He is a member of the ACM and the IEEE.
\end{IEEEbiography}

\end{document}

%% file: introduction.tex
\IEEEraisesectionheading{\section{Introduction}\label{sec:introduction}}

\subsection{Background and Motivation}\label{sec:MMQ_intro}

\IEEEPARstart{R}{eusing}
an appropriate materialized model (i.e., model after training) is proven helpful for a new learning task~\cite{DBLP:journals/tkde/PanY10}. It is called model reuse~\cite{DBLP:journals/pvldb/HasaniTAKD18, DBLP:conf/icde/Sigl19} in the database community and transfer learning~\cite{DBLP:conf/nips/NeyshaburSZ20} in the ML community, and it is widely used in many data-driven applications~\cite{DBLP:conf/icde/FangYZZ16, DBLP:conf/kdd/HanHAB21}. Model reuse$\footnote{\footnotesize We use model reuse and transfer learning interchangeably in this work.}$ leverages the knowledge of the materialized model (i.e., initial model) to improve the learning performance (e.g., high accuracy, low loss)~\cite{DBLP:journals/tkde/XuPXWLMS17, DBLP:journals/pvldb/JinSWDK21}.
Traditional model reuse problem assumes that the materialized model is given. However, the assumption is impractical in real-world applications since the developers need to find a proper materialized model first~\cite{DBLP:conf/ijcai/XiangPPSY11, DBLP:conf/icde/MiaoLDD17, DBLP:journals/pr/AfridiRS18}. 
Plenty of ML models are materialized (i.e., trained on their own tasks and stored) in model repositories~\cite{DBLP:journals/pvldb/HasaniTAKD18, DBLP:conf/icde/Sigl19} and managed in ML model management systems~\cite{DBLP:conf/icde/MiaoLDD17, DBLP:conf/sigmod/VartakSLVHMZ16} by  researchers and companies since the extensive usage of ML and model reuse. 
Nevertheless, an inappropriate materialized model may incur low performance and even inversely hurt the target performance (i.e., negative transfer~\cite{DBLP:journals/tkde/PanY10}), which is against the purpose of model reuse.
Thus, a significant problem is how to find appropriate materialized models from the model repositories when given a new learning task.

Materialized model query methods aim to solve the problem by automatically finding the most appropriate materialized model for a new learning task. A brute-force way is to retrain all the candidate models, which is obviously unbearable. It needs nearly two weeks to retrain 10 ResNet-50 models 100 epochs on the ImageNet-2012 dataset with a 3090 GPU. Under this circumstance, it is essential to find appropriate materialized models efficiently and accurately~\cite{DBLP:conf/icde/Sigl19, DBLP:journals/pr/AfridiRS18}, as in the following cases we try to support.

\vspace{0.05in}
\noindent\textbf{Case 1: (Big Company).}
With the high demand for ML in commercial products, big companies (e.g., Google) train and store plenty of ML models for different tasks or applications. Besides, training one ML model may generate multiple models since deep learning needs to repetitively adjust the models, such as hyper-parameter tuning. Even if the models are trained from the same data, they may lead to different results in model reuse. In addition, the ML models will be updated as the application version changes. As a result, a big company can easily store thousands of ML models in its materialized model repository. In this case, given a new task, it is difficult to find an appropriate materialized model for reuse via a brute-force way.   

\vspace{0.05in}
\noindent\textbf{Case 2: (Individual Developer).}
With the rapid development of the ML community, it is easy for individual developers to download materialized models from the online repositories, e.g., Github, TensorFlow Hub and PyTorch Hub, etc. Furthermore, individual developers generally do not have enough GPU cards, which makes the brute-force way much more time-consuming than companies.

\vspace{0.05in}
\noindent\textbf{Case 3: (One-stop AI Open Cloud Platform).}
One-stop AI open cloud platforms, such as Azure ML and Cloud AutoML,
provide one-stop training service for the datasets given by users. These platforms often use model reuse to improve the final performance. Thus, it is important for them to query appropriate materialized models efficiently and accurately for model reuse.

In these situations, it is significant to reduce the query time while still giving helpful answers. Motivated by this, materialized model query aims to find top-$k$ appropriate materialized models from a materialized model repository for a given target task. In the ML community, materialized model query is also called source model selection~\cite{DBLP:journals/pr/AfridiRS18, DBLP:conf/ijcai/XiangPPSY11}.

\begin{figure}[t]
\centering
\vspace{-2mm}
\includegraphics[width=0.48\textwidth]{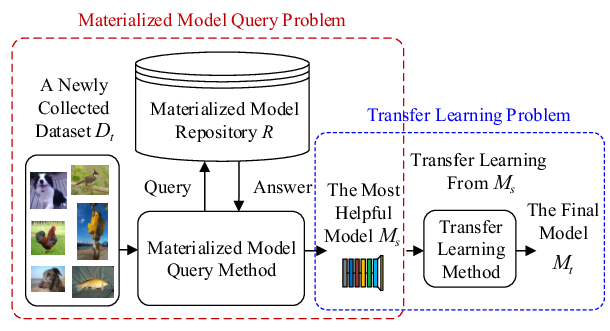}
\vspace{-4mm}
\caption{Illustration of Materialized Model Query}
\vspace{-7mm}
\label{fig:framework_little}
\end{figure}

\vspace{0.05in}
\noindent\textbf{General Workflow of Materialized Model Query.}
As shown in Fig.~\ref{fig:framework_little}, a data scientist, Alice, gets a new image classification task on a newly collected image dataset $D_t$. She decides to reuse a materialized model from model repository $R$ to proceed transfer learning on $D_t$. Next, she conducts the materialized model query, and obtains an appropriate materialized model $M_s$ from $R$.
Finally, she uses a transfer learning method to transfer the knowledge from $M_s$ to her new task and get the final model $M_t$. If the materialized model query method works well, $M_t$ will have good performance on $D_t$.

\vspace{0.05in}
\noindent\textbf{The Difference between Materialized Model Query and Transfer Learning.}
It is worth noting that, materialized model query is not transfer learning but the precondition of transfer learning.
Materialized model query problem does not concern what transfer learning method Alice uses. It aims at finding appropriate materialized models from the model repository for all kinds of transfer learning methods, which is the left part of Fig.~\ref{fig:framework_little}; while transfer learning methods focus on improving the learning performance of a specified materialized model, which is the right part of Fig.~\ref{fig:framework_little}. With the same transfer learning method, a better materialized model query method finds a more appropriate materialized model $M_s$ from model repository $R$ to make the final model $M_t$ perform better on $D_t$.
Hence, without loss of generality, we use the general transfer learning method, i.e., retraining, in our experiments (Section~\ref{sec:experiment}).

\vspace{-2mm}
\subsection{Limitations of Existing Studies}
Recently, a lot of efforts have been devoted to materialized model query from many database~\cite{DBLP:conf/icde/Sigl19, DBLP:journals/pvldb/HasaniTAKD18} and ML~\cite{DBLP:conf/ijcai/XiangPPSY11, DBLP:journals/pr/AfridiRS18} researchers.
Besides, several model management systems~\cite{DBLP:conf/icde/MiaoLDD17, DBLP:conf/sigmod/00010DD17} also provide materialized model query methods to enable users to find appropriate materialized models for model reuse.
Unfortunately, each approach has some limitations, which incurs three major problems.
The limitations of state-of-the-art papers are shown in Table~\ref{tab:works}.

\vspace{0.05in}
\noindent{\textbf{1) Source Data Required.}
Several existing studies~\cite{DBLP:conf/icde/Sigl19, DBLP:conf/ijcai/XiangPPSY11, DBLP:conf/icde/MiaoLDD17, DBLP:conf/sigmod/00010DD17, DBLP:journals/pvldb/HasaniTAKD18} need the training data of materialized models (i.e., source data) to answer materialized model queries.
However, letting the model uploaders upload the materialized model along with the training data is time-consuming, as the training data is often relatively large in order to enable the materialized model to achieve better transfer learning results on the target task.
Besides, training data could be related to user privacy and company secrets.
Thus, most companies and developers usually share materialized models on Github without any source data. Some existing works~\cite{DBLP:conf/icde/Sigl19, DBLP:conf/ijcai/XiangPPSY11, DBLP:conf/icde/MiaoLDD17, DBLP:conf/sigmod/00010DD17} utilize the similarity between source data and target data to decide which materialized model is more appropriate to the target task. Unfortunately, the similarity between source data and target data cannot represent the relevance between materialized models and target task.
Previous studies have shown that the similarity between source data and target data does not completely determine the result of transfer learning~\cite{DBLP:journals/pr/AfridiRS18, DBLP:conf/nips/NeyshaburSZ20}. Actually, there are many factors that determine whether the materialized model has appropriate knowledge for the target task, such as the way of training (e.g., epoch, learning rate, etc.) and dataset characteristics (e.g., similarity between source data and target data, dataset size, data quality, etc.). 
}


\begin{table}[t]
\caption{Limitations of State-of-the-Arts}
\vspace{-3mm}
\centering
\renewcommand\arraystretch{1.1}
\begin{tabular}{|c|c|c|c|}
\hline
\textbf{Paper} & \textbf{{Source-data Free}} & \textbf{Generality} & \textbf{Efficiency} \\ \hline
Hasani et al.~\cite{DBLP:journals/pvldb/HasaniTAKD18} & No & Low & High \\ \hline
Sigl~\cite{DBLP:conf/icde/Sigl19} & No & Low & High \\ \hline
Miao et al.~\cite{DBLP:conf/sigmod/00010DD17} & No & Low & High \\ \hline
Xiang et al.~\cite{DBLP:conf/ijcai/XiangPPSY11} & No & Low & High \\ \hline
Afridi et al.~\cite{DBLP:journals/pr/AfridiRS18} & Yes & Medium & Low \\ \hline
\textbf{Our Method} & \textbf{Yes} & \textbf{High} & \textbf{High} \\ \hline
\end{tabular}
\label{tab:works}
\vspace{-8mm}
\end{table}

\vspace{0.05in}
\noindent\textbf{2) Lack of Generality.}
Several existing materialized model query methods only work when the labels of materialized models are the same as~\cite{DBLP:conf/icde/Sigl19, DBLP:journals/pvldb/HasaniTAKD18} or relate to~\cite{DBLP:conf/ijcai/XiangPPSY11, DBLP:conf/sigmod/00010DD17} the labels of the target dataset.
However, it is common to transfer the knowledge from a materialized model trained on a dataset that is greatly different from the target dataset in transfer learning~\cite{DBLP:journals/nature/EstevaKNKSBT17, kermany2018identifying, DBLP:conf/nips/NeyshaburSZ20}.
Back to the real-world example, materialized models are usually from very different domains.
For instance, due to small data and expensive labeling for medical image datasets, the models trained on traditional image datasets (e.g., ImageNet-2012 dataset) are widely used in medical image classification task, and can achieve higher performance than training from scratch~\cite{kermany2018identifying, DBLP:journals/nature/EstevaKNKSBT17}.
In this scenario, labels (e.g., goldfish, ostrich, baseball, etc.) of the source dataset (i.e., ImageNet-2012) are quite different from those (e.g., choroidal neovascularization, diabetic macular edema, etc.) of the target dataset (i.e., a medical image dataset), which makes it difficult to select appropriate materialized models for the target task.
In addition, many types of model structures (e.g., CNN, RNN, SVM, etc.) and machine learning tasks (e.g., image classification, text classification, regression, etc.) exist. However, some previous studies are designed for specific models or tasks. SSFTL~\cite{DBLP:conf/ijcai/XiangPPSY11} can be employed to find the materialized model only for the text classification task.
The CNN automatic source model selection framework (CAS)~\cite{DBLP:journals/pr/AfridiRS18} can only be used in the classification task and cannot simultaneously deal with various model structures. In addition, the framework relies on the operation of the network structure, which makes it only work on specific model structures.
It is worth mentioning that CAS is with medium generality since CAS is the only existing work that can automatically find materialized models when the source labels are quite different from the target labels and work without any source data.


\vspace{0.05in}
\noindent\textbf{3) Inefficiency.}
The model training process of machine learning (especially deep learning) is time-consuming.
Although the previous work~\cite{DBLP:journals/pr/AfridiRS18} tries to reduce the training time, it still needs a training process, which is not efficient.

Overall, the existing materialized model query approaches are not powerful enough for general materialized model query or inefficient.
{Motivated by these, we propose \textsf{MMQ}, a source-data free, general, efficient, and effective materialized model query framework that works without source data and training process.}

\vspace{-2mm}
\subsection{Contributions and Organization}
\noindent
\textbf{Contributions.}
In this paper, we solve the materialized model query problem from a data management standpoint. 
{
We focus on fine-tuning since it is one of the most popular transfer learning method for deep learning models. Fine-tuning is widely used and studied in various areas and has achieved good results~\cite{DBLP:conf/emnlp/MouMYLX0J16, DBLP:conf/nips/FromeCSBDRM13, DBLP:conf/cvpr/0002ZV022, DBLP:journals/sensors/YamanakaNHIM22, DBLP:journals/tnn/RoCHC22}. However, this simple transfer learning method is limited by the relationship between the source model and the target task, and the selection of an inappropriate source model will have a negative effect on the transfer effectiveness. Thus, MMQ aims to provide an effective method to quickly select the source model for this common transfer learning method, i.e., fine-tuning.}
Since the purpose of materialized model query is to find the most appropriate materialized models for a given target task, we can naturally think of constructing a metric, from the perspective of data management. 
The metric should mainly answer two questions: (i) how to express the relevance between the materialized models and the given target task, and (ii) how to efficiently compute the metric and rank materialized models.

We propose a novel metric called separation degree, which solves the above two questions well.
Separation degree measures target-related knowledge of materialized models, i.e., how much knowledge the model contains that can help the target task. Then we select the top-$k$ materialized models with the most target-related knowledge.
Specifically, given a target task, \textsf{MMQ} directly applies the materialized model on the target dataset without any training, which outputs a representation vector for each sample in the target dataset. The representation vector is further normalized by the extended softmax function, and becomes a suitable probability vector to indicate the probability of the sample being each class. We utilize the Gaussian distribution to fit for the probability vector distribution of each class, based on which the separation degree of Gaussian distributions is introduced as the metric to measure the target-related knowledge of the materialized model. At last, \textsf{MMQ} ranks the materialized models based on separation degree, and selects the top-$k$ models with the highest separation degree as the answer to the query.

{To sum up, (i) since \textsf{MMQ} works without source data of materialized models, the framework is \textbf{privacy-protected}; (ii) our metric computation avoids high training cost, and thus, it is \textbf{efficient}; (iii) our metric is able to measure the relevance between the materialized model and the given target task, and hence, it is \textbf{effective}; and (iv) our metric calculation does not rely on any specific model structure and only uses the labels of target dataset, and thus, the metric is \textbf{general} to support various types of AI models and tasks. Also, we further improve the efficiency of our materialized model query method by proposing \textsf{I-MMQ}.}

The key contributions are summarized as follows:
\begin{itemize}[leftmargin=*]
\vspace{-0.5mm}
\item \textit{Efficient materialized model query framework.}
We present an efficient and general materialized model query framework \textsf{MMQ} for model reuse. To the best of our knowledge, \textsf{MMQ} is the first general framework to find appropriate materialized models without any training process and any source data.
\textsf{MMQ} is able to i) work without any source data; ii) support the cases when the source labels are completely different from the target labels; iii) support any type of AI models; and iv) offer efficient computation.
\item \textit{Metric for target-related knowledge.} We use the materialized model to embed samples in the target dataset into probability vectors, based on which we design a Gaussian mixture-based metric, i.e., separation degree, to evaluate the target-related knowledge of the materialized model on the target dataset.
\item \textit{Improved MMQ.} We propose the improved \textsf{MMQ}, i.e., \textsf{I-MMQ}, which utilizes an effective dimensionality reduction method to cut down the model query time, and meanwhile, select appropriate materialized models.
\item \textit{Extensive experiments.} We conduct extensive experiments on six public datasets and three types of tasks to verify the effectiveness and efficiency of the proposed approaches on a range of practical model reuse workloads.
\end{itemize}

\noindent
\textbf{Outline.}
The rest of this paper is organized as follows.
Section~\ref{sec:preliminaries} introduces basic preliminary materials related to the materialized model query, and defines the problem of materialized model query.
Section~\ref{sec:framework} presents the overall architecture of our proposed \textsf{MMQ}, and Section~\ref{sec:alg} details the proposed method and the improved method.
Section~\ref{sec:experiment} reports our experimental results.
Section~\ref{sec:relatedwork} reviews the related work.
Finally, we conclude the paper in Section~\ref{sec:conclusions}. 

%% file: preliminaries.tex
\vspace{-1mm}
\section{Preliminaries}
\label{sec:preliminaries}

\begin{table}[t]
\caption{Symbols and Description}
\centering
\vspace{-3mm}
\begin{tabular}{|l|l|}
\hline
\textbf{Notation} & \textbf{Description} \\ \hline
$M_i$  & \begin{tabular}[c]{@{}l@{}}a materialized model trained on the source\\ dataset $D_i$\end{tabular}\\ \hline
$D_t$  & a target dataset\\ \hline
$M_{(i,t)}$  & \begin{tabular}[c]{@{}l@{}}the final model after transfer learning  on\\  $D_t$ from $M_i$\end{tabular}\\ \hline
$(x_l, y_l)\in D_t$  & a sample belonging to the dataset $D_t$\\ \hline
$Z^i_l = (z_1, ..., z_n)$ & \begin{tabular}[c]{@{}l@{}}the output vector of $M_i$ with $x_l$ as input\end{tabular}\\ \hline
$P$ & \begin{tabular}[c]{@{}l@{}}the number of output units of a materialized \\model in the process\end{tabular}\\ \hline
$m$ & the number of classes in target dataset $D_t$\\ \hline
\end{tabular}
\label{tab:notation}
\vspace{-5mm}
\end{table}

\begin{figure*}[t]
	\centering
	\vspace{-2mm}
	\includegraphics[width=1\textwidth]{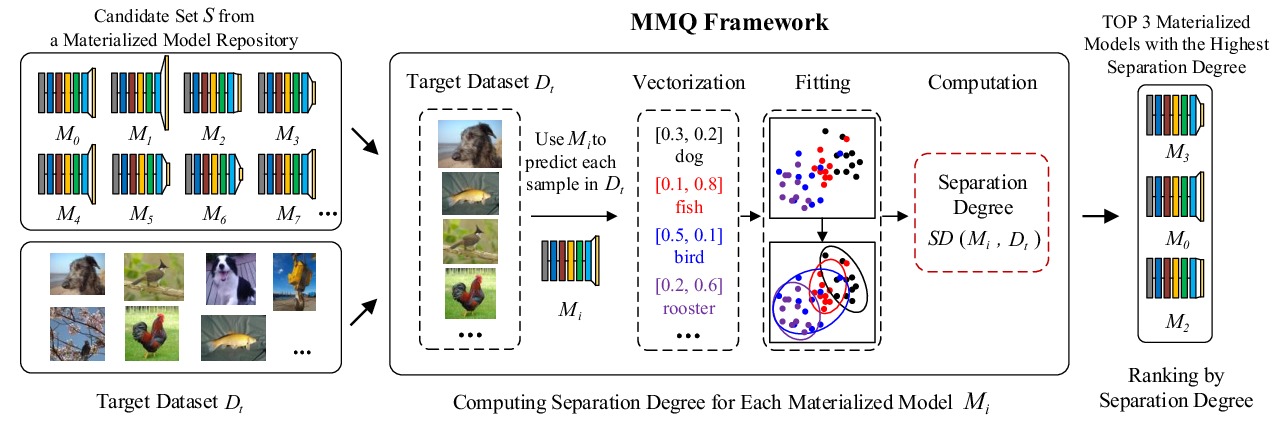}
	\vspace{-6mm}
	\caption{An Overview of \textsf{MMQ} Framework}
 	\vspace{-5mm}
	\label{fig:overview}
\end{figure*}

In this section, we first present some background materials to be used in subsequent sections and then formalize the problem of the materialized model query. Table~\ref{tab:notation} summarizes the notations frequently used throughout this paper.

\vspace{-1mm}
\subsection{Materialized Model Reuse}\label{sec:problem_def}
Materialized model reuse has the same meaning as transfer learning, and is always used in the database community since materialization and reuse are two fundamental concepts in database optimization~\cite{DBLP:journals/tkde/AksuCCKU14, DBLP:conf/sigmod/AgrawalCN01, DBLP:journals/pvldb/AtzeniBPT19, DBLP:conf/sigmod/DursunBCK17}. Hence, we introduce the concepts of materialized model and model reuse.

\vspace{0.05in}
\noindent\textbf{Materialized Model.} $M_i$ is the materialized model trained on source dataset $D_i$. For different materialized models $M_i$ and $M_j$, $D_i$ and $D_j$ can be completely different. For example, $D_i$ contains pictures of dogs and cats, while $D_j$ contains pictures of different traffic lights.

\vspace{0.05in}
\noindent\textbf{Transfer Learning (Model Reuse).} Given a materialized model $M_i$ and a target dataset $D_t = \{(x_l, y_l), l = 1, ..., N \}$, transfer learning methods leverage the knowledge from $M_i$ to improve the performance of the final model $M_{(i,t)}$ on $D_t$ (e.g., high accuracy, low loss). Since materialized model query methods aim at finding appropriate materialized models from the model repository for all kinds of transfer learning methods, we utilize a general transfer learning method, i.e., retraining, for experiments. In addition, transfer learning is also called model reuse.

\vspace{0.05in}
\noindent\textbf{Retraining.} In real-world applications, a target dataset  $D_t$ is always divided into a training set $D_{t_\alpha}$ and a test set $D_{t_\beta}$. Given a materialized model $M_{i}$ and a target dataset $D_t = \{D_{t_\alpha}, D_{t_\beta}\}$, retraining means that we use the materialized model $M_{i}$ as the initial parameters, and train $M_{i}$ on $D_{t_\alpha}$ to get the final model $M_{(i,t)}$.

\vspace{0.05in}
\noindent\textbf{Performance.} Given a materialized model $M_{i}$ and a target dataset $D_t = \{D_{t_\alpha}, D_{t_\beta}\}$, the performance of $M_{i}$ after retraining is the result quality (e.g., accuracy, loss) of the final model $M_{(i,t)}$ on $D_{t_\beta}$.

\vspace{-0.05in}
\subsection{Top-$k$ Materialized Model Query Problem}\label{sec:problem_def}
Based on the aforementioned preliminaries, we formally define the top-$k$ materialized model query problem.

\vspace{2mm}
\begin{definition}
Given a materialized model candidate set $S = \{M_{1}, ..., M_{n}\}$ and a target dataset $D_t=\{D_{t_\alpha}, D_{t_\beta}\}$, the top-$k$ materialized model query aims to find $k$ materialized models with the best retraining performance.
\end{definition}

\noindent\textbf{Ground Truth of The Problem.} The retraining performance is only used as the ground truth in experiments.
Generally, materialized model query methods neither retrain any of the materialized models nor rank the models by their performance after retraining.
Instead, materialized model query methods rank all the materialized models in order via the proposed metrics since the retraining is too time-consuming. The ranking results based on actual retraining performance are only used as the ground truth in experiments for testing.
Therefore, the better the materialized model query method, the closer the ranking results are to the ground truth. 

%% file: framework.tex
\vspace{-2mm}
\section{Framework Overview}\label{sec:framework}
In this section, we first overview the framework of \textsf{MMQ}, and then, we describe each step of \textsf{MMQ} framework.

Fig.~\ref{fig:overview} depicts the overview of the \textsf{MMQ} framework. \textsf{MMQ} takes as inputs a candidate set $S$ from a materialized model repository and a target dataset $D_t$, and computes the ranking metric (i.e., the separation degree) of each materialized model on $D_t$.
Specifically, each materialized model $M_{i}$ in $S$ is processed by the following procedures:

\begin{enumerate}[leftmargin=*]
\item \textbf{Vectorization.} First, \textsf{MMQ} directly applys $M_{i}$ on the target dataset $D_{t}$ with a extended softmax function, and thus, all samples in $D_{t}$ are transformed into probability vectors (i.e., soft labels) (to be detailed in Section~\ref{sub:vector}).

\item \textbf{Fitting.} Second, \textsf{MMQ} uses multivariate Gaussian distributions to fit for clusters of sample vectors (to be detailed in Section~\ref{sub:fitting}).

\item \textbf{Computation.} Third, \textsf{MMQ} computes the separation degree of $M_{i}$ based on Gaussian distributions (to be detailed in Section~\ref{sub:computation}).
\end{enumerate}

By ranking the separation degree of each materialized model, the top-$k$ materialized models can be selected.

%% file: techniques.tex
\vspace{-2mm}
\section{The Proposed \textsf{MMQ}}
\label{sec:alg}

In this section, we describe the proposed \textsf{MMQ}, an efficient materialized model query framework for general model reuse.
We first investigate how to construct the metric for the materialized model query, and then detail three phases of the \textsf{MMQ} framework.
Next, we present how to extend \textsf{MMQ} to regression task.
Finally, we provide an improved \textsf{MMQ} to further reduce the query time of \textsf{MMQ}.

\vspace{-2mm}
\subsection{Metric}
\label{sub:motivation}

The purpose of model reuse (i.e., transfer learning) is to use the knowledge which is related to target task and is possessed by the materialized model (i.e., \textbf{target-related knowledge}) to assist the training of the final model. A materialized model with more target-related knowledge is likely to achieve higher performance after model reuse.
Existing studies~\cite{DBLP:conf/icde/Sigl19, DBLP:conf/ijcai/XiangPPSY11, DBLP:conf/sigmod/00010DD17} try to capture the target-related knowledge by measuring the similarity between the source dataset of the materialized model and the target dataset. Nonetheless, there are many factors that determine the target-related knowledge of a materialized model, such as the way of training (e.g., epoch, learning rate, etc.) and dataset characteristics (e.g., similarity between source and target datasets, dataset size, data quality, etc.). Using the similarity between source and target datasets is not able to describe all of the factors.
To this end, we decide to directly measure the target-related knowledge of materialized models via a newly designed metric.

\begin{figure}[tb]
	\centering
	\vspace{-1mm}
	\hspace{-0.25cm}
	\subfigure[A random model]{
		\includegraphics[width=0.215\textwidth]{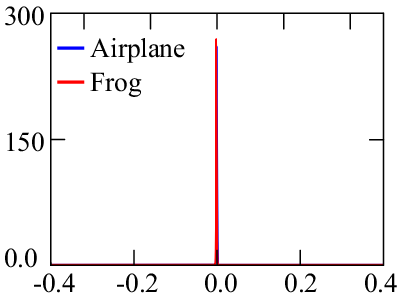}}
	\hspace{1.4mm}
	\hspace{-0.25cm}
	\subfigure[A trained model]{
		\includegraphics[width=0.215\textwidth]{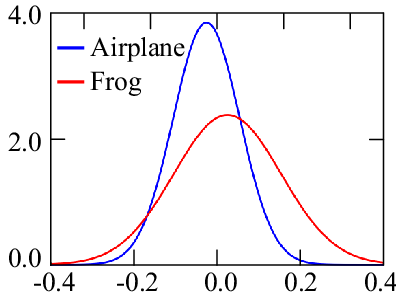}}
	\vspace{-3mm}
	\caption{Separation Degree of Two Materialized Models}
	\vspace{-5mm}
	\label{fig:introduction_2}
\end{figure}

We propose a new metric called \textbf{separation degree} to measure the target-related knowledge of materialized models, which bases on a property of feature extractor~\cite{DBLP:journals/pr/AfridiRS18, lda, DBLP:conf/icml/ChenK0H20}, i.e., a good feature extractor with more knowledge for a dataset can better separate different classes of samples in this dataset after feature extraction.
Besides, deep learning models can be regarded as feature extractors, and the prediction process of deep learning models can also be regarded as feature extraction.
Therefore, a materialized model with more target-related knowledge can better separate different classes of samples in the target dataset after the prediction of this materialized model.

To be more specific, we compared the separation degree of two LeNet-like CNN~\cite{lecun1998gradient} materialized models. The first model is built with random parameters (without training), while the second model is trained on two classes of FashionMNIST dataset (i.e., Coat and Bag). As confirmed by the experiments, the second model achieves better performance than the first model after retraining, i.e., the second model has more target-related knowledge.
Here, for each materialized model, we use it to predict each sample of two classes from CIFAR-10 dataset (i.e., Airplane and Frog) with a softmax function, and get the output probability vector for each sample. Then, we construct two Gaussian distributions of those probability vectors with Airplane and Frog labels separately. Finally, we present the Gaussian distributions generated by each materialized model respectively in Fig.~\ref{fig:introduction_2}. The blue curves represent Airplane, and the red curves denote Frog.
As shown in Fig.~\ref{fig:introduction_2}, all the vectors predicted by the random model with less target-related knowledge are crowded together (Fig.~\ref{fig:introduction_2}(a)). In contrast, the trained model with more target-related knowledge has better ability to separate different classes of vectors in the target dataset (Fig.~\ref{fig:introduction_2}(b)).
Hence, if a materialized model has higher separation degree on the target dataset, it means that the model contains more target-related knowledge, and has better performance after model reuse.

Thus, we compute and rank the separation degree of materialized models in our proposed \textsf{MMQ}. As shown in Fig.~\ref{fig:overview}, \textsf{MMQ} consists of three phases, i.e., (i) vectorization, (ii) fitting, and (iii) separation degree computation.

\vspace{-1mm}
\subsection{Vectorization}
\label{sub:vector}
Given a materialized model candidate set $S = \{M_{1}, ..., M_{q}\}$ and a target dataset $D_t = \{(x_l, y_l), l = 1, ..., N \}$. For each materialized model $M_{i}$ in $S$, we first vectorize all the training samples in the target dataset according to $M_{i}$, i.e., we directly utilize $M_{i}$ to predict each sample $(x_l, y_l) \in D_t$, and get a probability vector, called \textbf{soft label}. Here, $x_l$ is the training data, $y_l$ is the corresponding label of $x_l$, and $N$ is the number of samples in $D_t$.

Specifically, without any training, we predict each sample $x_l$ using $M_{i}$ to get the hidden representation vector $Z = (z_1, ..., z_n)$ from the final layer. Next, $Z$ is further normalized by the extended softmax function to get the soft label $x^i_l$ below:

\begin{equation}
x^i_l= (\alpha_1, ..., \alpha_n),  \text{as } \alpha_m = \frac{\exp \left(z_{m} / T\right)}{\sum_{j = 1}^{n} \exp \left(z_{j} / T\right)} (1 \le m \le n).
\end{equation}

$T$ is a parameter in the extended softmax function. Note that, the probability vector is called soft label since it is the vector that denotes the probability distribution over classes. In other words, the materialized model gives a possible "label" for each sample in the target dataset.
For example, consider a materialized model trained to identify dogs and cats. If we directly apply it to the target dataset, it may generate a soft label (0.9, 0.1) for a tiger picture, and a soft label (0.49, 0.51) for an airplane picture. The first soft label indicates that the tiger picture is 90\% possibility of being a cat and 10\% possibility of being a dog, while the second soft label means that the airplane picture is 49\% possibility of being a cat and 51\% possibility of being a dog.

The extended softmax function becomes the normal softmax function when the parameter $T$ is set to 1.
The larger the $T$ is, the smaller the difference among probabilities will be, i.e., a larger $T$ makes a softer probability distribution over classes~\cite{DBLP:conf/cvpr/YuanTLWF20}.
For instance, when $T$ is set to 1, the soft label for a sample is {$(0.09, 0.83, 0.08)$}; and when $T$ is set to 5, the soft label is {$(0.29, 0.44, 0.27)$}.
We use the extended softmax function rather than the normal softmax function because a softer probability distribution from $T$ larger than 1 produces better measurement results of target-related knowledge, as confirmed in Section~\ref{ImpactofT}.

After embedding each sample $x_l$ into a soft label vector $x^i_l$, we delete one extra dimension in the soft label to avoid redundant computation. It leads to no information loss since the sum of elements in the soft label is 1.0 (due to the normalization of the extended softmax function). Finally, we get the soft label set $D^i_t = \{(x^i_l, y_l), l = 1, ..., N \}$.

\vspace{-1mm}
\subsection{Fitting}\label{sub:fitting}

Next, we divide the soft labels $D^i_t$ into different clusters according to their actual labels in the target dataset $D_t$. Assume that the number of labels in the target dataset $D_t$ is $m$. We can obtain $m$ clusters $\{C_1, ..., C_m\}$, where $C_1  \cup \cdots  \cup  C_m = D^i_t$, and each cluster $C_u (1 \le u \le m)$ contains soft labels belonging to one label. Note that, the actual label is different from soft label. To be more clear, for each sample $(x_l, y_l) \in D_t$, the soft label is generated from the training data $x_l$ by $M_{i}$, while the actual label is $y_l$ which is corresponding to $x_l$.

Our model query framework is general. It aims at various machine learning models and tasks. Hence, it is difficult to make specific assumptions about the data distribution of soft labels. Gaussian distribution is the most widely used distribution~\cite{DBLP:journals/tkde/LuCL21, DBLP:journals/tkde/XieQZPC22, DBLP:journals/tkde/HeCSBH11}. According to the generalization of the central limit theorem~\cite{clt}, a considerable number of phenomena in nature produce a final distribution that is approximately normal. In view of this, we assume that soft labels of each cluster follow Gaussian distribution, which is verified its effectiveness in our experiments (see Section~\ref{sec:exp_performance}).
Multivariate dimensional Gaussian distribution generalizes the one-dimensional Gaussian distribution to higher dimensions. We use it since most soft labels are vectors with more than one dimension. For simplification, "multivariate" is omitted in the subsequent contents.

Thus, we fit for each cluster $C_u$ in $\{C_1, ..., C_m\}$ using a multivariate dimensional Gaussian distribution $G_u$.
\begin{equation}
G_u({x}) = \frac{\exp \left(-\frac{1}{2}({x}-{\mu_u})^{\mathrm{T}} {\Sigma_u}^{-1}({x}-{\mu_u})\right)}{\sqrt{(2 \pi)^{n}|{\Sigma_u}|}} ,
\end{equation}
where $\Sigma_u$ denotes the covariance matrix of $C_u$ and $\mu_u$ represents the mean vector of $C_u$.

\subsection{Computation of Separation Degree}\label{sub:computation}
In the sequel, we define the separation degree between two Gaussian distributions, based on which, we also define the separation degree of the materialized model.

\vspace{2mm}
\begin{definition}\label{defn: Separation degree between two clusters}
{\bf (Separation degree between two Gaussian distributions)}
\textit{
Given two multivariate dimensional Gaussian distributions $G_u({x})$ and $G_v({x})$ fitted for two clusters $C_u$ and $C_v$, respectively. The peak of $G_u({x})$ and $G_v({x})$ are $P_u = G_u(\mu_u)$ and $P_v = G_v(\mu_v)$, respectively. A mixture of Gaussian distribution $F({x}) = G_u({x}) + G_v({x})$. The separation degree between $G_u({x})$ and $G_v({x})$ is defined as:
\begin{equation}
SD_{(u,v)} = SD_{(v,u)} = \frac{P_u}{F(\mu_u)} + \frac{P_v}{F(\mu_v)} - 1.
\end{equation}
{Here, $D_{(u,v)} \in [0, 1)$. The larger $D_{(u,v)}$ is, the more separated between two Gaussian distributions $G_u(\cdot)$ and $G_v(\cdot)$.}
After changing the form, and setting
\begin{equation}
e_u({x})=\exp \left(-\frac{1}{2}({x}-{\mu_u})^{\mathrm{T}} {\Sigma_u}^{-1}({x}-{\mu_u})\right),
\end{equation}
the formula of separation degree can be obtained as follows:
\begin{equation}
\label{equ:comp}
\frac{\sqrt{| \Sigma_{v} |} }{\sqrt{| \Sigma_{v}|}+\sqrt{| \Sigma_{u}|} e_{v}(\mu_u)}
+ \frac{\sqrt{| \Sigma_{u} |}}{\sqrt{| \Sigma_{v}|}e_{u}(\mu_v)+\sqrt{| \Sigma_{u}|}}
- 1.
\end{equation}}
\end{definition}

The formula of separation degree between two Gaussian distributions after the form changing can be computed efficiently. Besides, the form of determination (i.e., $|\Sigma_{u}|$ and $|\Sigma_{v}|$) division can avoid the problem where the determination value is too large or too small to express.

\begin{figure}[tb]
	\centering
	\hspace{-0.25cm}
	\includegraphics[width=0.295\textwidth]{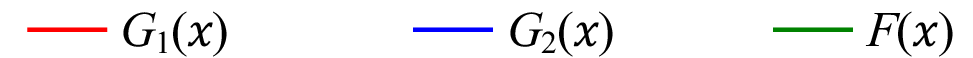}

	\vspace{2mm}
	\subfigure[$SD_{(1,2)} = 1.0$ ]{
		\includegraphics[width=0.155\textwidth]{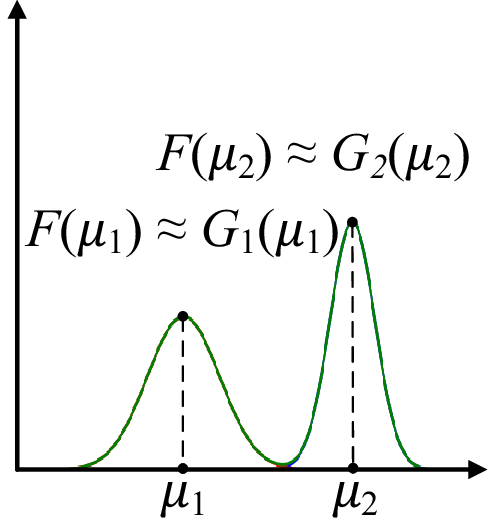}}
	\hspace{-0.4cm}
	\subfigure[$SD_{(1,2)} = 0.5$ ]{
		\includegraphics[width=0.155\textwidth]{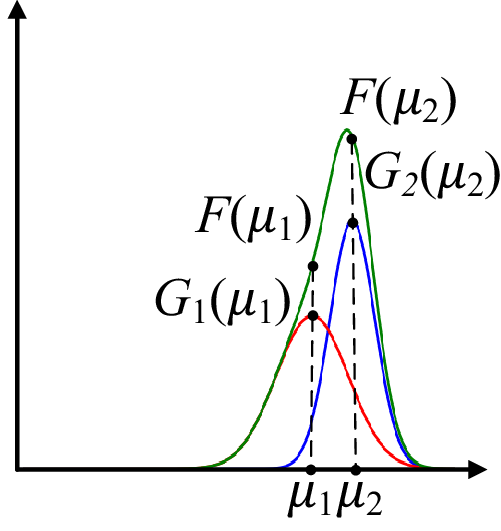}}
	\hspace{-0.4cm}
	\subfigure[$SD_{(1,2)} = 0.0$]{
		\includegraphics[width=0.155\textwidth]{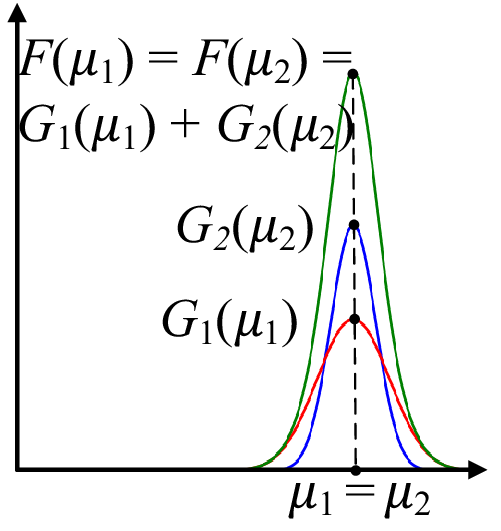}}
	\vspace{-3mm}
	\caption{Separation Degree between Gaussian Distributions}
	\vspace{-5mm}
	\label{fig:gaussian}
\end{figure}

The reason behind Definition~\ref{defn: Separation degree between two clusters} is that $P_u(\mu_u)/F(\mu_u) \in [0.5, 1)$ and $P_v(\mu_v)/F(\mu_v) \in [0.5, 1)$ are both monotonically decreasing when two separated Gaussian distributions $G_u$ and $G_v$ gradually approach until their center points coincide ($\mu_u = \mu_v$). Thus, Definition~\ref{defn: Separation degree between two clusters} is able to measure the degree of separation of two Gaussian distributions.

Fig.~\ref{fig:gaussian} illustrates three different cases of the separation degree between two Gaussian distributions. Here, $\mu_1$ and $\mu_2$ denote the mean value of $G_1({x})$ and $G_2({x})$, respectively. As $\mu_1$ is getting closer to $\mu_2$, $SD_{(1,2)}$ is changing from 1.0 to 0.0. Fig.~\ref{fig:gaussian}(a) shows the situation when $G_1({x})$ and $G_2({x})$ are far apart, in which $F({\mu_1}) \approx G_1({\mu_1})$, $F({\mu_2}) \approx G_1({\mu_2})$, and $SD_{(1,2)} = SD_{(2,1)} \approx 1.0$. Fig.~\ref{fig:gaussian}(b) depicts the situation when $G_1({x})$ and $G_2({x})$ are close, where $SD_{(1,2)} = SD_{(2,1)} = 0.5$. Fig.~\ref{fig:gaussian}(c) plots the situation when $G_1({x})$ and $G_2({x})$ are the closest such that $\mu_1 = \mu_2$, in which $F({\mu_1}) = F({\mu_2}) = G_1({\mu_1}) + G_2({\mu_2})$ and $SD_{(1,2)} = SD_{(2,1)} = 0.0$.

Since different classes of samples are usually not highly related, we calculate the separation degree of a materialized model using the average separation degree between any two Gaussian distributions. Thus, we define the separation degree of the materialized model as follows.

\begin{algorithm}[t]
    \small
    \LinesNumbered
    \label{alg:1}
    \caption{Materialized Model Query (\textsf{MMQ})}
    \KwIn{a candidate set $S = \{M_{1}, ..., M_{q}\}$, a target dataset $D_t = \{(x_l, y_l), l = 1, ..., N \}$, parameter $T$, integer $k$}
    \KwOut{top-$k$ appropriate materialized models}
    \For{each $M_{i}$ in $S$}
	{
	    $C_{all} \leftarrow  \varnothing$; \tcp{the list to store clusters}
	    $G_{all} \leftarrow  \varnothing$; \tcp{the list to store Gaussian distributions}
	    $SD_{all} \leftarrow  \varnothing$; \tcp{the list to store the separation degree between every two  Gaussian distributions}
        \For{each $(x_l, y_l)$ in $D_t$} {
            use $M_{i}$ to predict $x_l$, and get the hidden representation vector $Z = (z_1, ..., z_n)$ from the final layer;\\
            use the extended softmax function with parameter $T$ to normalize $Z$, and get the soft label $x^i_l$; \tcp{Eq.~1}
            delete the last dimension in the soft label $x^i_l$;\\
            $p \leftarrow y_l$;\\
            \If{$C_p$ not in $C_{all}$}{
                \tcp{{Check if $C_p$ has been created}}
                $C_p \leftarrow  \varnothing$; \tcp{$C_p$ is the cluster of label $p$}
                $C_{all} \leftarrow C_{all} \cup \{C_p\}$;
    		}
            get $C_p$ from $C_{all}$;\\
            $C_p \leftarrow C_p \cup \{x^i_l\}$;\\
            update $C_p$ in $C_{all}$;
        }
        \For{each cluster $C_u$ in $C_{all}$} {
            fit a multivariate dimensional Gaussian distribution $G_u$ to cluster $C_u$; \\
            $G_{all} \leftarrow G_{all} \cup \{G_u\}$;
        }
        \For{each $G_u$ in $G_{all}$} {
            \For{each $G_v$ in $G_{all}$} {
                calculate the separation degree $SD_{(u,v)}$ between $G_u$ and $G_v$; \tcp{Eq.~5}
            }
        }
        calculate the separation degree $SD\left(M_{i}, D_{t}\right)$ of materialized model $M_{i}$ by averaging $SD_{all}$; \tcp{Eq.~6}
    }
    \Return{$k$ materialized models with the $k$ highest separation degrees}
\end{algorithm}

\begin{definition}\label{defn: Separation degree of materialized model}
	{\bf (Separation degree of a materialized model)}
	\textit{
	Given a candidate set $S$, a target dataset $D_t$,
	{and the number $m$ of label types in the target dataset $D_t$, }
	the separation degree of materialized model $M_{i} \in S$ on $D_t$ is defined as:
	\begin{equation}
	SD\left(M_{i}, D_{t}\right)=\frac{1}{m^2} \sum_{C_{u} \subseteq D^i_{t}} \sum_{C_{v} \subseteq D^i_{t}} SD_{(u, v)}.
    \end{equation}
	}
\end{definition}

We use the separation degree to rank the materialized models on the target dataset. Finally, $k$ materialized models having the $k$ highest separation degrees are the top-$k$ best models we select for the query.

The pseudo-code of the materialized model query is depicted in Algorithm~\ref{alg:1}. \textsf{MMQ} takes as inputs a candidate set $S$, a target dataset $D_t$, a parameter $T$, and an integer $k$, and outputs top-$k$ appropriate materialized models.
Note that, each data sample in the target dataset $D_t$ contains the data $x_l$ and corresponding label $y_l$.
For each materialized model $M_{i}$ in the candidate set $S$, the algorithm first generates three lists to store clusters, Gaussian distributions, and separation degrees separately (lines 1--4).
Thereafter, for each data sample $(x_l, y_l)$ in the target dataset, \textsf{MMQ} uses $M_{i}$ to predict $x_l$, and gets the hidden representation vector $Z = (z_1, ..., z_n)$ from the final layer of $M_{i}$ (lines 5--6).
Then, it uses the extended softmax function with parameter $T$ to normalize $Z$ to the soft label $x^i_l$ via Eq.~1, and delete the last dimension with no information loss (lines 7--8).
{
Next, we use label $p$ to represent label $y_l$ of sample $(x_l, y_l)$ (line 9), and cluster soft labels according to the labels of their corresponding samples, i.e., we put the soft labels $x^i_l$ having label $p$ into the cluster $C_p$ for further fitting process (lines 10--15). 
Specifically, for soft label $x^i_l$ generated by $(x_l, p)$ ($p$ equals to $y_l$), we first check whether the cluster $C_p$ exits, and initialize it as an empty set if it does not exist (lines 10--12). Subsequently, we add the soft label $x^i_l$ in $C_p$ (lines 13--15).}
In the sequel, for each cluster $C_u$,  it fits a multivariate dimensional Gaussian distribution $G_u$ to $C_u$ and inserts $G_u$ into $G_{all}$ (lines 16--18). Then, it calculates the separation degree $SD_{(u,v)}$ between every two Gaussian distributions via Eq.~5, and computes the separation degree $SD\left(M_{i}, D_{t}\right)$ of materialized model $M_{i}$ by averaging $SD_{all}$ via Eq.~6 (lines 19--22). Finally, $k$ materialized models with the $k$ highest separation degrees are returned (line 23).

\subsection{Extend \textsf{MMQ} to Regression Task}
\textsf{MMQ} is also flexible to support the regression task because it only ranks materialized models by measuring their target-related knowledge.

The separation degree for regression task is different from that of classification task since there is an additional relationship between samples and labels in regression task. Specifically, the samples whose labels are 0 are similar to those whose labels are 1, but far from the samples whose labels are 100.
For example, a photo of a 100-year-old person is more like that of a 99-year-old person than that of a 20-year-old person.
Note that, sample labels are original data labels that are different from soft labels in \textsf{MMQ}.
Therefore, samples with similar labels are similar to a certain extent, while samples with dissimilar labels are different. Hence, \textsf{MMQ} tries to use the weight $|u-v|^p$ to keep this relationship, where $u$ and $v$ are discretized labels after we discrete the continuous value labels of the target dataset.
By adding this weight, two Gaussian distributions with low separation degree but far away in terms of their sample labels will be punished, which follows the rule of regression task. Thus, we define the separation degree of the materialized model for regression task as follows.

\vspace{2mm}
\begin{definition}\label{defn: Separation degree of materialized model in regression tasks}
	{\bf (Separation degree of materialized model for regression task)}
	\textit{
	Given a candidate set $S$, a target dataset $D_t$, and a norm parameter $p \geq 0$,
	the separation degree of materialized model $M_{i} \in S$ for regression task on $D_t$ is defined as follows:
	\begin{equation}
	SD\left(M_{i}, D_{t}\right)=\frac{\sum_{C_{u} \subseteq D^i_{t}} \sum_{C_{v} \subseteq D^i_{t}} \left|u-v\right|^p SD_{(u, v)}}{\sum_{C_{u} \subseteq D^i_{t}} \sum_{C_{v} \subseteq D^i_{t}} \left|u-v\right|^p}.
    \end{equation}
	}
\end{definition}

Note that, given two discretized class labels $u$ and $v$ in the target dataset $D_t$, if $u$ and $v$ are far from each other in the original continuous value, the weight $\left|u-v\right|^p$ would be large in order to emphasize the importance to distinguish $u$ and $v$.

\vspace{1mm}
\subsection{Improved \textsf{MMQ}}

\begin{figure}[t]
\centering
\includegraphics[width=0.44\textwidth]{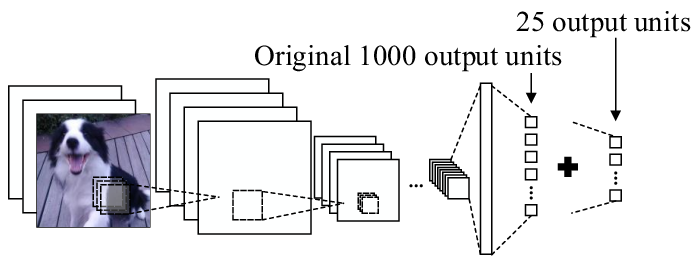}
\vspace{-4mm}
\caption{Illustration of \textsf{I-MMQ}}
\vspace{-7mm}
\label{fig:I-EMS}
\end{figure}

Although \textsf{MMQ} can efficiently find appropriate materialized models without any training process, it still needs to compute the separation degree of each materialized model, which contains two expensive operations, i.e., the matrix determination and the matrix inverse in Eq.~\ref{equ:comp}. Let $P$ be the dimensionality of soft labels generated by a materialized model (i.e., the number of output units), the time complexities of the two operations are both $O(P^3)$, which is costly, especially when $P$ is large.

In order to alleviate this problem, we further propose an Improved \textsf{MMQ} (\textsf{I-MMQ} for short), which utilizes a dimensionality reduction method to cut down $P$. 
{The design of I-MMQ aims to improve the efficiency of MMQ, which requires an efficient dimensionality reduction method while ensuring that the ranking of materialized models based on separation degree is not significantly affected. 
Since MMQ calculates the overlap between Gaussians, manifold learning methods that are designed to preserve the data distribution structure and distance relationships can maintain the overlap relations between Guassians after dimensionality reduction to some extent.
However, these dimensionality reduction methods, e.g., LLE~\cite{roweis2000nonlinear}, Isomap~\cite{balasubramanian2002isomap}, are time-consuming. 
For faster dimensionality reduction, we consider a simple way of dimensionality reduction, i.e., we propose \textsf{I-MMQ}, which adds a fully connected layer with fewer output units and random parameters to the final output of the materialized model (without using the activation function). 
Fig.~\ref{fig:I-EMS} illustrates the process of \textsf{I-MMQ}. A materialized model is trained on ImageNet-2012, which has 1000 different classes. Thus, it is costly for \textsf{MMQ} to compute the materialized model's separation degree on the target dataset since there are 1000 output units.
Although the stochastic linear dimensionality reduction performed by adding a fully connected layer could affect the distance relationship between soft labels, it may not lead to significant changes in the computation of the separation degree.
This is because MMQ ranks the feature extraction capability of models by the degree of overlapping (or separation degree) of Gaussian distributions fitted from soft labels rather than relying entirely on the distance relationship, i.e., inter-class distance and intra-class distance. 
As verified in Section 5.2.8, the use of Gaussian separation to calculate the feature extraction ability of the model is better than the cluster evaluation metrics using Inter-class distance and intra-class distance. 
It is difficult to judge how much the impact of dimensionality reduction has on computing the degree of Gaussian separation compared with manifold learning methods.
Thus, we conduct experiments (in Section~\ref{sec:Reduction}) to compare I-MMQ with the manifold learning methods
, which proves that I-MMQ with a fully connected layer can achieve good dimensionality reduction performance and is much more efficient than the rest of the methods.
}


\vspace{0.05in}
\noindent{\textbf{Computational complexity.} 
In order to study the efficiency of MMQ and I-MMQ, we analyze the computation time complexity of our two methods. Let $P$ be the dimensionality of soft labels  (i.e., the number of output units), $m$ be the number of class in the target dataset, $N$ be the data quantity of the target dataset. Besides, we take two steps, i.e., fitting and computation, to calculate the separation degree from soft labels. The complexity of fitting step is $O(NP^2)$ for calculating covariance matrix, while the time complexity of computation step is $O(m^2P^3)$. Thus, the total time complexity of MMQ is 
\begin{equation}
O(NP^2+m^2P^3).
\end{equation}
I-MMQ reduces the dimensionality of soft labels to a constant number (i.e., a much smaller value). Thus, the total time complexity of I-MMQ is 
\begin{equation}
O(N+m^2).
\end{equation}
Since the predicting time depends on model structure, we analyze the time complexity of MMQ and I-MMQ without vectorization step, and we also study the running time except predicting in the subsequent experiments.}

%% file: experiment.tex
\section{Experiments}
\label{sec:experiment}

In this section, we conduct extensive experiments to evaluate the effectiveness and efficiency of our proposed \textsf{MMQ} and \textsf{I-MMQ} compared with the state-of-the-art competitors.

\vspace{-3mm}
\subsection{Experimental Setup}\label{sub:setup}

\textbf{Datasets.}
We employ three types of tasks, i.e., image classification, text classification, and image regression. Nine public datasets were used. Specifically, ImageNet-2012$\footnote{\url{http://www.image-net.org/challenges/LSVRC/2012/2012-downloads}}$ contains 1,331,167 images from 1000 classes;
{Places365-Standard$\footnote{\url{http://places2.csail.mit.edu/}}$ contains 1,800,000 images from 344 classes;}
{CUB-200-2011$\footnote{\url{https://data.caltech.edu/records/65de6-vp158}}$ contains 11,788 images from 200 classes;}
{CIFAR-100$\footnote{\url{https://www.cs.toronto.edu/~kriz/cifar.html}}$ contains 60,000 images from 100 classes;}
CIFAR-10$\footnote{\url{https://www.cs.toronto.edu/~kriz/cifar.html}}$ includes 60,000 images from 10 classes; 
FashionMNIST$\footnote{\url{ https://github.com/zalandoresearch/fashion-mnist}}$ contains 60,000 images from 10 classes; 
IMDB$\footnote{\url{https://www.kaggle.com/datasets/lakshmi25npathi/imdb-dataset-of-50k-movie-reviews}}$ contains 50,000 movie reviews from 2 classes;
RCV1$\footnote{\url{http://riejohnson.com/cnn_data.html}}$ includes 15,564 news from 55 classes;
and House Price$\footnote{\url{https://github.com/emanhamed/Houses-dataset}}$ contains 2,140 images from 535 sample houses in California, USA.
As shown in Table~\ref{tab:E0}, based on the six datasets, we support five cases.
\begin{itemize}[leftmargin=*]
\item \textbf{Case 1}: \textbf{Image classification task} using different parts of ImageNet-2012 as the source datasets and the target dataset separately;
\item \textbf{Case 2}: \textbf{Image classification task} using FashionMNIST as the source datasets and CIFAR-10 as the target dataset;
\item \textbf{Case 3}: \textbf{Text classification task} using IMDB as the source datasets and RCV1 as the target dataset;
\item \textbf{Case 4}: \textbf{Image regression task} using CIFAR-10 as the source datasets and House Price as the target dataset;
{
\item \textbf{Case 5}: \textbf{Image classification task} using 3 different datasets as the source datasets and CIFAR-100 as the target dataset.}
\end{itemize}

For each case in the experiments, we construct a candidate set by generating materialized models from the source datasets, and get the ground truth performance by retraining on the target dataset.
\textbf{Note that, the actual retraining performance is only used as the ground truth.} The materialized model query methods only use their own metrics to rank materialized models.
Therefore, the better the materialized model query method, the closer the ranking results are to the ground truth.
In addition, the labels of the target dataset are different from those of the source datasets in each experimental case for verifying generality.

Specifically, for \textbf{Case 1}, we divide ImageNet-2012 dataset into two sub-datasets for transfer learning.
One sub-dataset $D_s$ contains 700 classes, and the other sub-dataset $D_t$ includes the rest 300 classes as target dataset. 
{Thus, the data quantity of the target dataset in Case 1 is 191,739, and the number of classes in the target dataset is 300.}
The labels of $D_s$ and $D_t$ are completely different. We use $D_s$ to randomly generate 56 different sub-datasets, where the number of classes in each sub-dataset ranges from 2 to 200.
To construct the candidate set, 56 ResNet-18~\cite{DBLP:conf/cvpr/HeZRS16} materialized models are trained on 56 sub-datasets, respectively, and 50-90\% data of each sub-dataset is randomly used for training. We set the batch size to 512 and the number of training epochs to 20. In addition, ImageNet-2012 pre-trained model provided by PyTorch-torchvision$\footnote{\url{https://pytorch.org/vision/stable/index.html}}$ and the untrained ResNet-18 model are also considered as materialized model candidates. The output unit number of the two models are 1000 and 300, respectively. Finally, We use the above 58 materialized models as the candidate set.
To get the ground truth accuracy, we retrain each model in candidate set using the stochastic gradient descent with Nesterov momentum for optimization, and set the initial learning rate to 0.1, the momentum to 0.9, the batch size to 512, and the weight decay to $1 \times 10^{-4}$, and the learning rate is divided by 10 every 30 training epochs. We take Cross-entropy as the loss function.
Besides, we use early stopping as the training stop condition, and if the loss of validation set does not decrease in 10 epochs, we shut down the retraining, and take the model of the lowest loss. The number of output units in \textsf{I-MMQ} is set to 25.
The parameter $T$ is set to 2.0 as default.

\begin{table}[]
\caption{Information of 4 Cases}
\vspace{-3mm}
\label{tab:E0}
\small
\begin{tabular}{c|c|c|c}
\hline
\rule{0pt}{8pt} \textbf{Case} & \textbf{Source Data}   & \textbf{Target Data} & \textbf{Task Type}   \\ \hline
Case 1        & \begin{tabular}[c]{c}ImageNet-\\2012\end{tabular} & \begin{tabular}[c]{c}ImageNet-\\2012\end{tabular}    & \begin{tabular}[c]{c}Image\\Classification\end{tabular} \\ \hline
Case 2        & \begin{tabular}[c]{c}Fashion-\\MNIST\end{tabular}  & CIFAR-10      & \begin{tabular}[c]{@{}c@{}}Image\\Classification\end{tabular} \\ \hline
Case 3        & IMDB          & RCV1       & \begin{tabular}[c]{@{}c@{}}Text\\Classification\end{tabular}  \\ \hline
Case 4        & CIFAR-10      & House Price      & \begin{tabular}[c]{@{}c@{}}Image\\Regression\end{tabular}     \\ \hline
{Case 5}        & \begin{tabular}[c]{@{}l@{}}{\romannumeral1.~ImageNet-}\\ 2012\\ \romannumeral2.~Places\\ \romannumeral3.~CUB-200-\\ 2011\end{tabular}      & CIFAR-100      & \begin{tabular}[c]{@{}c@{}}Image\\Regression\end{tabular}     \\ \hline
\end{tabular}
\vspace{-6mm}
\end{table}

For \textbf{Case 2}, we first randomly generate 200 different sub-datasets from FashionMNIST. The number of classes in each sub-dataset ranges from 2 to 5.
To construct the candidate set, 200 models of LeNet-like CNN are trained on the 200 sub-datasets, respectively. 1-99\% data of each sub-dataset is randomly used for training, and we set the batch size to 64 and training epochs to 20.
We take a subset of CIFAR-10 with four classes as the target dataset.
{Thus, the data quantity of the target dataset in Case 2 is 24,000, and the number of classes in the target dataset is 4.}
To get the ground truth accuracy, we retrain each materialized model using Adam, and set the base learning rate to $3 \times 10^{-4}$ and the batch size to 64. Besides, we use Cross-entropy as the loss function. Like Case 1, we also use early stopping.

For \textbf{Case 3}, 50 materialized models of RNN with 128 hidden size and LSTM units are trained on IMDB respectively to construct the candidate set. 1-99\% data of IMDB is randomly used for training. We set the batch size to 64 and the number of training epochs to 20. We use the GloVe.6B.100d vectors as the pre-trained word embeddings.
We take RCV1 with 55 classes as the target dataset.
{Thus, the data quantity of the target dataset in Case 3 is 15,564, and the number of classes in the target dataset is 55.}
To get the ground truth accuracy, we use Adam for retraining, and set the base learning rate to $3 \times 10^{-4}$ and the batch size to 128. The early stopping is also used.

For \textbf{Case 4}, we firstly generate 50 VGG-11~\cite{DBLP:journals/corr/SimonyanZ14a} network models on the subsets of the CIFAR-10 dataset to construct the candidate set, where the number of classes in each sub-dataset changes from 2 to 10, and 50-90\% data of each sub-dataset is randomly used for training.
The 50 trained materialized models constitute the candidate set.
We take House Price dataset as the target dataset.
{Thus, the data quantity of the target dataset in Case 4 is 2,140.}
To get the ground truth loss, we also use Adam, and set the base learning rate to 0.001 and the batch size to 32. We take mean absolute percentage error as the loss function, and use the early stopping. In addition, we discretize the continuous labels into 10 discrete labels with the same number of samples.

{
For \textbf{Case 5}, we use 12 different structures, i.e., Resnet-34~\cite{DBLP:conf/cvpr/HeZRS16}, Resnet-50, Resnet-101, iResnet-34~\cite{DBLP:conf/icpr/DutaL0020}, iResnet-50, iResnet-101, VGG-11~\cite{DBLP:journals/corr/SimonyanZ14a}, VGG-13, VGG-16, VGG-19, Googlenet~\cite{googlenet} and Densenet121~\cite{densenet}, and 3 source datasets, i.e., ImageNet-2012, Places, and CUB-200-2011, for generating 108 materialized models.
Specifically, we generate 3 materialized models on the subsets of each source dataset for each structure. The sub-datasets are constructed by randomly sampling 10-90\% classes and 50-90\% data from the source dataset.
We use CIFAR-100 as the target dataset. Thus, the data quantity of the target dataset in Case 5 is 60,000, and the number of classes in the target dataset is 100.
To get the ground truth accuracy, we use the same retraining setting as Case 1. The early stopping is also used, and the number of output units in \textsf{I-MMQ} is set to 25.}

\vspace{0.1in}
\noindent\textbf{Competitors.}
We compare the proposed \textsf{MMQ} and \textsf{I-MMQ} with two existing materialized model query frameworks: CNN source model selection framework (CAS)~\cite{DBLP:journals/pr/AfridiRS18} and REAPER~\cite{DBLP:conf/icde/Sigl19}.
To the best of our knowledge, CAS is the only existing work that can automatically find materialized models when the source labels are quite different from the target labels and works without source data, like the proposed \textsf{MMQ} and \textsf{I-MMQ}. Thus, we mainly compare \textsf{MMQ} with CAS.
REAPER is the latest materialized model query method, and it uses the similarity between the source dataset and the target dataset to query materialized models. {However, it lacks generality and needs source data.} REAPER only works when all the source datasets are given, and source labels are exactly the same as the target labels, which is different from the goal of our \textsf{MMQ}. Therefore, we only compare \textsf{MMQ} with REAPER in Section~\ref{sub:metadata}. Since the specific implementation of REAPER is not explained in \cite{DBLP:conf/icde/Sigl19}, we use meta-data to measure the similarity between the source dataset and the target dataset.

\vspace{0.1in}
\noindent\textbf{Evaluation metrics.}
To evaluate the effectiveness of materialized model query, we use three different metrics.
(i) \textit{Pearson Correlation Coefficient $($PCC$)$.} PCC is utilized to measure the correlation between the ranking metric and the accuracy of the materialized model retrained on the target dataset precisely (i.e., the ground truth accuracy). Here, the ranking metric is the normalized separation degree for \textsf{MMQ} (or \textsf{I-MMQ}), the normalized ranking score for CAS, and the normalized KLD and JSD for REAPER. The larger the PCC, the better the performance of the materialized model query framework is.
(ii) \textit{Lowest Accuracy of Top-$k$ Ranked Materialized Models.} In order to measure the materialized model query ability of different frameworks, we compare the lowest accuracy of the top-$k$ materialized models ranked by different frameworks.
(iii) \textit{Trendline.} Trendline is fitted by the least squares method. It represents the trend of points, and reflects the correlation between the ranking metric and the accuracy after retraining to a certain extent.

To verify the efficiency of materialized model query, we divide the running time into two parts, i.e., the predicting time and the other time. The predicting time corresponds to the cost where the frameworks use each materialized model to predict each target sample directly, and output the hidden representation vector. The other time corresponds to the cost where the frameworks measure the target-related knowledge of each materialized model via their own metrics.

All experiments were implemented in Python 3.6.12 and PyTorch 1.6.0, and run on a Dell server with Ubuntu 18.04.5 LTS, Intel Xeon Silver 4210R 2.40GHz CPU, 128GB RAM, and GeForce RTX 2080Ti GPU. The source codes of \textsf{MMQ} are available online\footnote{\url{ https://github.com/ZJU-DAILY/MMQ}}

\subsection{Performance Study}
\label{sec:exp_performance}
We conduct seven sets of experiments to demonstrate the effectiveness and efficiency of our proposed \textsf{MMQ} and \textsf{I-MMQ} compared against CAS and REAPER.

\subsubsection{Performance on ImageNet-2012}
\label{sec:ImageNet}

In the first set of experiments, we compare \textsf{MMQ} and \textsf{I-MMQ} with CAS on Case 1, where different subsets of ImageNet-2012 are used as the source datasets and the target dataset.

\begin{table}[]
\centering
\caption{Comparison of Performance and Efficiency on Case 1.}
\label{tab:E1}
\vspace{-3mm}
\renewcommand\arraystretch{1.1}
\begin{tabular}{c|c|c|c|c}
\hline
\multirow{2}{*}{\textbf{Datasets}} & \multirow{2}{*}{\textbf{Framework}} & \multirow{2}{*}{\textbf{\quad PCC \quad}} & \multicolumn{2}{c}{\textbf{Time (h)}} \\ \cline{4-5}
                                   &                                     &                              & \textbf{Predicting}  & \textbf{\quad Other\quad} \\ \hline
\multirow{3}{*}{\begin{tabular}[c]{@{}c@{}}Case 1\end{tabular}}          & \textsf{MMQ}                              & \textbf{0.791}                                                                                           & \multirow{3}{*}{2.347}  & 2.606              \\ \cline{2-3} \cline{5-5}
& \textsf{I-MMQ}                               & 0.638                                                                                           &                      & \textbf{0.648}              \\ \cline{2-3} \cline{5-5}
& CAS                                 & 0.549                                                                                           &                      & 7.650              \\ \hline
\end{tabular}
\vspace{-5mm}
\end{table}

\begin{figure}[tb]
	\centering
	\subfigure[\textsf{MMQ}]{
		\includegraphics[width=0.21\textwidth]{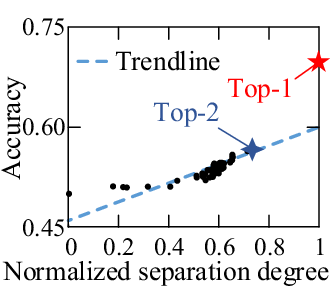}
	}
	\hspace{1.4mm}
	\hspace{-0.25cm}
	\subfigure[\textsf{I-MMQ}]{
		\includegraphics[width=0.21\textwidth]{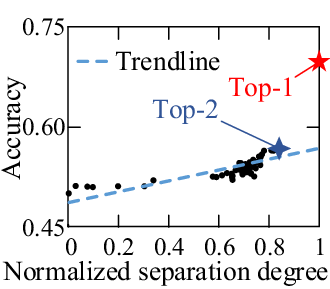}
	}
	\vspace{-4mm}
	
	\centering
	\hspace{-0.25cm}
	\subfigure[CAS]{
		\includegraphics[width=0.21\textwidth]{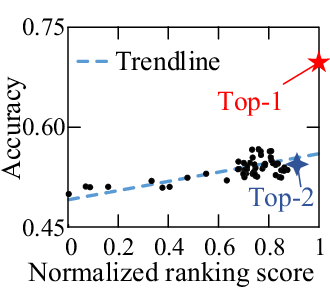}
	}
	\hspace{1.4mm}
	\hspace{-0.25cm}
	\subfigure[Lowest accuracy]{
		\includegraphics[width=0.21\textwidth]{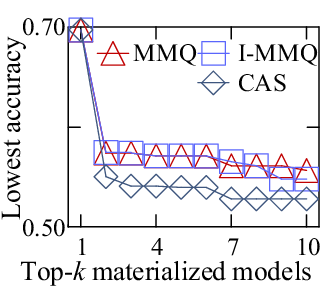}
	}
	\vspace{-3mm}
	\caption{Ranking Results on ImageNet-2012}
	\vspace{-6mm}
	\label{fig:E1}
\end{figure}

Table~\ref{tab:E1} depicts the PCC and running time of three different frameworks on Case 1.
First, we can observe that the PCC of \textsf{MMQ} and \textsf{I-MMQ} is higher than that of CAS.
This implies that the correlation between the ground truth accuracy and the ranking metric in \textsf{MMQ} and \textsf{I-MMQ} is better than that in CAS.
Second, all three different frameworks have the same predicting time, as they all predict each sample on the target dataset.
However, \textsf{MMQ} and \textsf{I-MMQ} have fewer other time costs than CAS.
As expected, \textsf{I-MMQ} is more efficient than \textsf{MMQ} as \textsf{I-MMQ} further reduces the number of output units to improve efficiency. 
Though the predicting time takes a high proportion of the total running time of \textsf{I-MMQ}, it can be reduced proportionally by sampling since the predicting time is linearly related to the number of samples. Therefore, we can further reduce the running time of \textsf{I-MMQ} by random sampling.

Figs.~\ref{fig:E1}(a) to \ref{fig:E1}(c) plot the accuracy of each materialized model retrained on the target dataset w.r.t. the ranking metric.
It is observed that all three frameworks, \textsf{MMQ}, \textsf{I-MMQ}, and CAS, can correctly rank the ImageNet-2012 pre-trained model as the top-1 model, which confirms the effectiveness of our methods. However, \textsf{MMQ} and \textsf{I-MMQ} can select the top-2 model correctly, while CAS selects a low-accuracy model as the top-2 materialized model.
Besides, the slope of the trendlines for \textsf{MMQ} and \textsf{I-MMQ} is larger than that for CAS, which validates the efficiency of our methods.
{
Fig.~\ref{fig:E1}(d) shows the lowest accuracy on the target dataset of top-$k$ materialized models.}
As observed, \textsf{MMQ} and \textsf{I-MMQ} can find better models than CAS when the performance of materialized models is close to each other.
In addition, the performance of \textsf{I-MMQ} is similar to that of \textsf{MMQ}, indicating that \textsf{I-MMQ} can achieve good performance even when the output dimension of materialized models is reduced to 25.

\begin{figure}[tb]
	\centering
	\subfigure[\textsf{I-MMQ} vs sampling rate]{
		\includegraphics[width=0.21\textwidth]{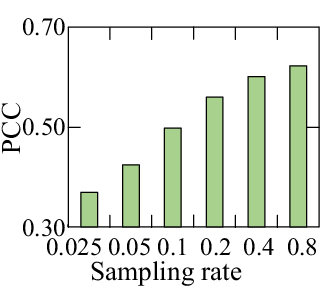}
	}
	\hspace{1.4mm}
	\hspace{-0.25cm}
	\subfigure[\textsf{I-MMQ} (sampling rate 0.1)]{
		\includegraphics[width=0.21\textwidth]{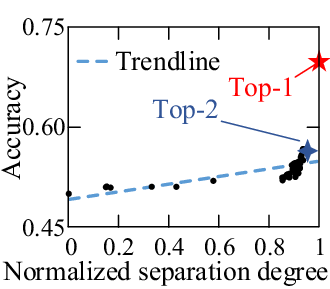}
	}
	\vspace{-2mm}
	\caption{Impact of Random Sampling on \textsf{I-MMQ}}
	\vspace{-4mm}
	\label{fig:E3}
\end{figure}

Next, we verify the robustness of \textsf{I-MMQ} on random sampling. Fig.~\ref{fig:E3}(a) shows the PCC of \textsf{I-MMQ} when the sampling rate of the target dataset varies from 0.025 to 0.8. It is observed that the PCC of \textsf{I-MMQ} increases with the sampling rate. This is because \textsf{I-MMQ} can better estimate the target-related knowledge of materialized model candidates as the sampling rate grows.
To be more specific, Fig.~\ref{fig:E3}(b) depicts the accuracy of materialized models retrained on the target dataset w.r.t. the normalized separation degree in \textsf{I-MMQ} when the sampling rate is 0.1.
We can observe that \textsf{I-MMQ} can select the top-2 models correctly, and perform better than CAS (shown in Fig.~\ref{fig:E1}(c)).
It confirms the robustness of \textsf{I-MMQ} on random sampling over the target dataset.
Thus, \textsf{I-MMQ} can reduce the predicting time by the random sampling on the target dataset while retaining the materialized model query performance.

Through the experiments on the large image classification dataset ImageNet-2012, we have shown that \textsf{I-MMQ} can enhance the efficiency and scalability of \textsf{MMQ} while maintaining the materialized model query performance. For simplification, we omit \textsf{I-MMQ} in the rest of the experiments due to similar experimental results.

\subsubsection{Performance on Different Datasets and Tasks}
The prior work~\cite{DBLP:conf/nips/NeyshaburSZ20} has shown that transfer learning can improve the performance of the target task even if the source dataset is quite different from the target dataset.
Thus, in this set of experiments, we compare \textsf{MMQ} with CAS in both image and text classification tasks when the source datasets and the target dataset are highly different {(i.e., using Case 2 and Case 3)}.

\begin{table}[]
\centering
\caption{Comparison of Performance and Efficiency on Case 2 and Case 3}
\vspace{-3mm}
\label{tab:E4-5}
\renewcommand\arraystretch{1.1}
\begin{tabular}{c|c|c|c|c}
\hline
\multirow{2}{*}{\textbf{Datasets}}                                                 & \multirow{2}{*}{\textbf{Framework}} & \multirow{2}{*}{\textbf{\quad PCC \quad}} & \multicolumn{2}{c}{\textbf{Time (min)}} \\ \cline{4-5}
                                                                                   &                                     &                              & \textbf{Predicting}  & \textbf{\quad Other\quad} \\ \hline
\multirow{2}{*}{\begin{tabular}[c]{@{}c@{}}Case 2\end{tabular}} & {\textsf{MMQ}}                               & \textbf{0.582}                                             & \multirow{2}{*}{0.159} & \textbf{1.446}          \\ \cline{2-3} \cline{5-5}
                                                                                   & CAS                                 & 0.032                                             &                        & 1.648          \\ \hline
\multirow{2}{*}{\begin{tabular}[c]{@{}c@{}}Case 3\end{tabular}}            & {\textsf{MMQ}}                               & \textbf{0.889}                                             & \multirow{2}{*}{0.489} & \textbf{1.341}          \\ \cline{2-3} \cline{5-5}
                                                                                   & CAS                                 & 0.526                                             &                        & 2.680          \\ \hline
\end{tabular}
\vspace{-5mm}
\end{table}

\begin{figure}[t]
	\centering
	\hspace{-0.25cm}
	\subfigure[Lowest accuracy (Case 2)]{
		\includegraphics[width=0.21\textwidth]{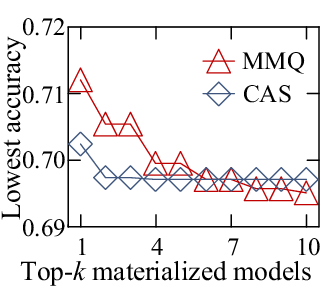}
	}
	\hspace{1.4mm}
	\hspace{-0.25cm}
	\subfigure[Lowest accuracy (Case 3)]{
		\includegraphics[width=0.21\textwidth]{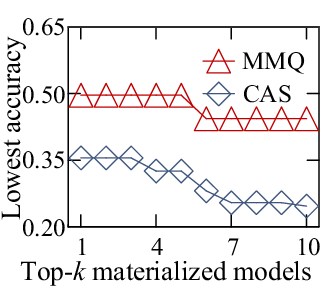}
	}
	\vspace{-2mm}
	\caption{{Ranking Results on Different Datasets and Tasks}}
	\vspace{-4mm}
	\label{fig:E4-5}
\end{figure}

Table~\ref{tab:E4-5} lists the PCC and running time of \textsf{MMQ} and CAS in the image classification task (i.e., FashionMNIST to CIFAR-10 datasets) and text classification task (i.e., IMDB to RCV1 datasets).
As observed, \textsf{MMQ} can get higher PCC than CAS in both of these tasks, meaning that the correlation between the ground truth accuracy and the ranking metric in \textsf{MMQ} is stronger than that of CAS for both image classification and text classification tasks.
Besides, \textsf{MMQ} can also get shorter running time than CAS, which again confirms the efficiency of \textsf{MMQ}.


Fig.~\ref{fig:E4-5}(a) and~\ref{fig:E4-5}(b) plot the lowest accuracy on the target dataset of the top-$k$ materialized models ranked by \textsf{MMQ} and CAS for the image classification and the text classification tasks, respectively.
It is observed from Fig.~\ref{fig:E4-5}(a) that \textsf{MMQ} first gives better top-$k$ query results and then becomes comparable with that of CAS as $k$ grows.
This is because the ground truth accuracy of candidate models is close to each other, resulting in close ranking results of \textsf{MMQ} and CAS.
Fig.~\ref{fig:E4-5}(b) shows that the lowest accuracy of \textsf{MMQ} is stably larger than that of CAS with the growth of $k$. This implies that \textsf{MMQ} can achieve better materialized model query performance than CAS in both the image classification and the text classification when the source datasets and the target dataset are greatly different.

\subsubsection{Performance on Regression Task}
\textsf{MMQ} is flexible to the regression task. In this set of experiments, we investigate the performance of \textsf{MMQ} on the regression task {for Case 4}.

\begin{figure}[t]
    \vspace{-3mm}
	\centering
	\hspace{-0.25cm}
	\subfigure[\textsf{MMQ} ($p=2$)]{
	\includegraphics[scale=0.52]{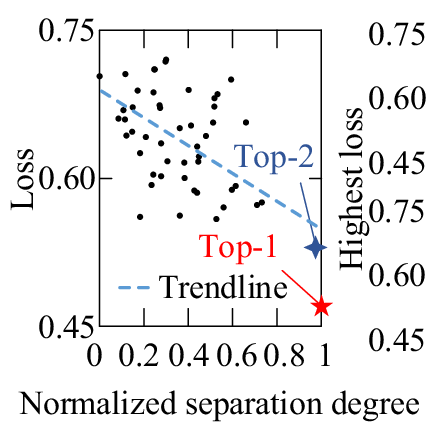}
	}
	\hspace{-3.5mm}
	\subfigure[{\textsf{MMQ} vs. $p$}]{
	\includegraphics[scale=0.52]{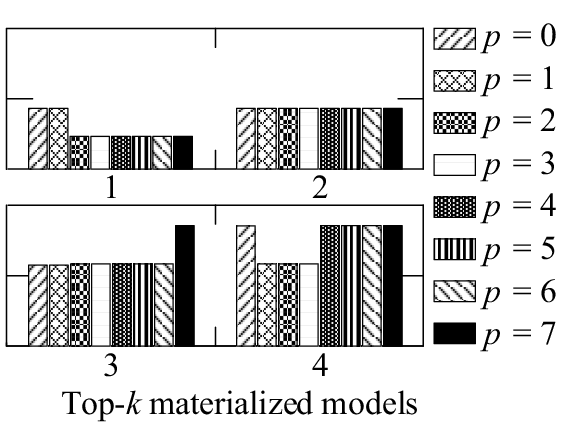}
	}
	\vspace{-2mm}
    \caption{\textsf{MMQ} Performance of Case 4}
	\label{fig:E6}
    \vspace{-6mm}
\end{figure}

Fig.~\ref{fig:E6}(a) depicts the loss of materialized models retrained on the target dataset w.r.t. the normalized separation degree computed by \textsf{MMQ} when the norm parameter $p = 2$.
It is observed that \textsf{MMQ} can correctly rank the top-2 models with the smallest loss after retraining on the target dataset. Besides, the slope of the trendline has a large absolute value. This validates the performance of \textsf{MMQ} when it extends to the regression task.
To further study the impact of $p$, we vary $p$ from 0 to 7.
Fig.~\ref{fig:E6}(b) plots the highest loss on the target dataset of top-$k$ materialized models selected by \textsf{MMQ}. We can observe that \textsf{MMQ} can achieve better materialized model query result when $p$ is 2, and as $p$ increases, the results are getting worse. The reason is that when $p$ is too large, \textsf{MMQ} will pay too much attention to the clusters having distant labels, and nearly completely ignore the clusters with closer labels, which is also wrong and harmful to the ranking results. Consequently, it would be better to set $p$ to 2 in practice.

\subsubsection{\textsf{MMQ} with Various Model Structures Simultaneously}

\begin{figure}[t]
\centering
\includegraphics[width=3.3in]{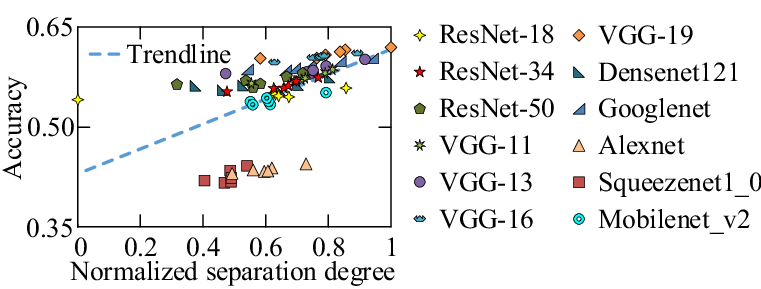}
\vspace{-3mm}
\caption{{\textsf{MMQ} Query Materialized Models of Different Structures}}
\vspace{-4mm}
\label{fig:E2.2}
\end{figure}

We further evaluate the robustness of \textsf{MMQ} using 12 commonly-used CNN networks on the ImageNet-2012 dataset simultaneously, which includes ResNet-18~\cite{DBLP:conf/cvpr/HeZRS16}, ResNet-34, ResNet-50, Densenet121~\cite{densenet}, VGG-11~\cite{DBLP:journals/corr/SimonyanZ14a}, VGG-13, VGG-16, VGG-19,  Googlenet~\cite{googlenet}, Alexnet~\cite{alexnet}, Squeezenet1\_0~\cite{squeezenet}, and Mobilenet\_v2~\cite{mobilenet}.
The experimental settings are the same as Case 1.
For each model structure, we train 6 materialized models on 6 sub-datasets, respectively. Thus, we generate 72 materialized models with 12 different structures as candidate materialized models.
Fig.~\ref{fig:E2.2} plots the ground truth accuracy of each materialized model w.r.t. normalized separating degree. It is observed that the correlation between our metric and the ground truth performance is strong. \textsf{MMQ} is able to find the best model having the highest accuracy on the target dataset, which confirms the robustness and generality of \textsf{MMQ} even for materialized models with different structures.

Fig.~\ref{fig:E2.2} also shows that the scatter points are distributed into two clusters, and the correlation between separation degree and the ground truth accuracy after retraining is strong within each cluster. Some models in the upper cluster have higher accuracy but lower separation degree than the models in the lower cluster.
The reason behind this is that the learning ability of Squeezenet1\_0 and Alexnet is much weaker than that of other CNN model structures.
It is known that ResNet-50 usually achieves higher accuracy than Alexnet if both are trained from scratch to converge on dataset $D_t$.
Similarly, even if Alexnet has more target-related knowledge than ResNet-50, the final accuracy of Alexnet may still be lower than ResNet-50 after model reuse.
As a result, there are two clusters in Fig.~\ref{fig:E2.2}. 
In general, we do not need to worry about this situation since researchers and developers usually consider materialized models of different structures with similar learning ability (i.e., ResNet and VGG). Thus, \textsf{MMQ} can return appropriate materialized models in general situations.

\begin{table}[]
\centering
\caption{{Average Running Time Except Predicting at Different Data Quantity of $D_t$ and Quantity of Category}}
\vspace{-3mm}
\label{tab:nE2}
\renewcommand\arraystretch{1.1}
\begin{tabular}{c|c|cccc}
\hline
\multirow{2}{*}{\textbf{Framework}}         & \multirow{2}{*}{\begin{tabular}[c]{@{}c@{}}\textbf{Changed}\\  \textbf{Rate}\end{tabular}} & \multicolumn{4}{c}{\textbf{Sample Rate}}                                                                     \\ \cline{3-6} 
                                &                                                                          & \multicolumn{1}{c|}{\textbf{20\%}}     & \multicolumn{1}{c|}{\textbf{40\%}}     & \multicolumn{1}{c|}{\textbf{60\%}}     & \textbf{80\%}     \\ \hline
\multirow{2}{*}{MMQ}            & class rate                                                               & \multicolumn{1}{c|}{4.764}   & \multicolumn{1}{c|}{17.413}  & \multicolumn{1}{c|}{38.839}  & 68.336  \\ \cline{2-6} 
                                & data rate                                                                & \multicolumn{1}{c|}{106.245} & \multicolumn{1}{c|}{107.137} & \multicolumn{1}{c|}{107.259} & 107.904 \\ \hline
\multirow{2}{*}{I-MMQ} & class rate                                                               & \multicolumn{1}{c|}{0.631}   & \multicolumn{1}{c|}{1.498}   & \multicolumn{1}{c|}{2.980}   & 4.924   \\ \cline{2-6} 
                                & data rate                                                                & \multicolumn{1}{c|}{7.475}   & \multicolumn{1}{c|}{7.491}   & \multicolumn{1}{c|}{7.501}   & 7.527   \\ \hline
\end{tabular}
\end{table}

\begin{figure}[t]
	\centering
	\vspace{-4mm}
	\subfigure[\textsf{MMQ}]{
		\includegraphics[width=0.21\textwidth]{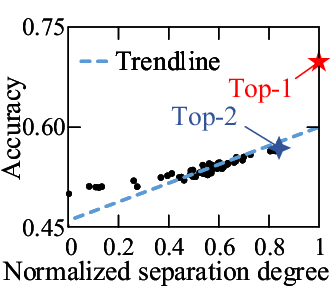}
	}
	\hspace{1.4mm}
	\hspace{-0.25cm}
	\subfigure[\textsf{I-MMQ}]{
		\includegraphics[width=0.21\textwidth]{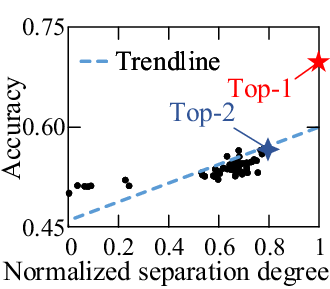}
	}
	\vspace{-4mm}
	\caption{{Ranking Results with 20\% Class Sampling Rate and 10\% Data Sampling Rate}}
	\vspace{-6mm}
\label{fig:nE2}
\end{figure}

\subsubsection{Impact of data quantity of $D_t$ and quantity of category}
{In order to better study the correlations between running time and the data quantity of the target dataset $D_t$, we report the running time of MMQ and I-MMQ by varying the data quantity and the number of classes in $D_t$ using Case 1. Specifically, we i) randomly sample 20\%, 40\%, 60\%, 80\% data from the target dataset to evaluate the effect of data quantity on the running time; and ii) randomly sample 20\%, 40\%, 60\%, 80\% classes from the target dataset while keeping the data quantity at 10\% of the total data to evaluate the effect of the number of target dataset classes on the running time. The corresponding results (i.e., the running time) have been reported in Table 6.}

{
As shown in Table~\ref{tab:nE2}, the computation time of MMQ and I-MMQ increases with the number of target dataset classes and the data quantity, which is consistent with the time complexity analysis. The second observation is that, the computation time of MMQ and I-MMQ increases faster with the number of classes compared with the data quantity, which is also consistent with the time complexity analysis. In addition, both MMQ and I-MMQ can achieve good ranking performance and select the best source model when the data and the classes are randomly sampled, as shown in Fig.~\ref{fig:nE2}. We randomly sample 20\% classes and keep the data quantity at 10\% of the total data, i.e., sample 50\% data in each class. Therefore, we can improve the efficiency of our methods through the random sampling without a large impact on the performance of source model selection.}

\begin{figure}[t]
	\centering
	\hspace{-0.25cm}
	\subfigure[MMQ]{
		\includegraphics[width=0.21\textwidth]{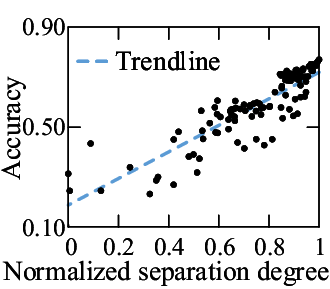}
	}
	\hspace{1.4mm}
	\hspace{-0.25cm}
	\subfigure[I-MMQ]{
		\includegraphics[width=0.21\textwidth]{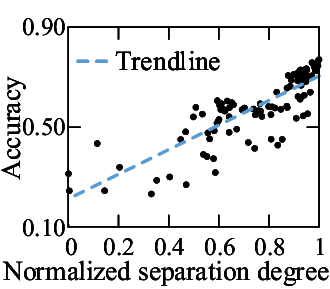}
	}
	\vspace{-2mm}
	\caption{{Ranking Results with Multiple Source Datasets and Multiple Model Structures Simultaneously on Case 5}}
	\label{fig:nE3}
\end{figure}

\subsubsection{\textsf{MMQ} with Various Datasets and Model Structures Simultaneously}
{In this set of experiments, we evaluate the performance of our \textsf{MMQ} and \textsf{I-MMQ} when there are multiple materialized models with different structures over multiple datasets in the candidate set {(i.e., using Case 5)}.}

{
Figs.~\ref{fig:nE3}(a) and \ref{fig:nE3}(b) plot the accuracy of each source model retrained on the target dataset w.r.t. normalized separating degree from MMQ and I-MMQ, respectively. It is observed that the correlation between our estimation metric (i.e., the normalized separation degree) and our methods (i.e., MMQ and I-MMQ) are able to select the best model with the highest accuracy. It confirms the effectiveness of our methods on source models of sufficiently different training data with different structures at the same time.}

\begin{figure}[t]
\vspace*{-4mm}
\centering
\hspace{-5mm}
\subfigure[Case 1]{
    \includegraphics[width=1.17in, height=1.12in]{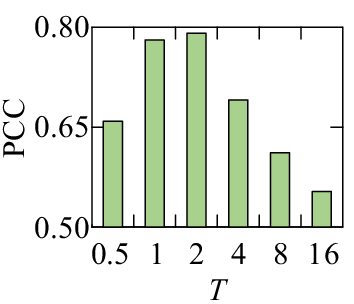}
}\hspace{-3.5mm}
\subfigure[Case 2]{
 \includegraphics[width=1.17in, height=1.12in]{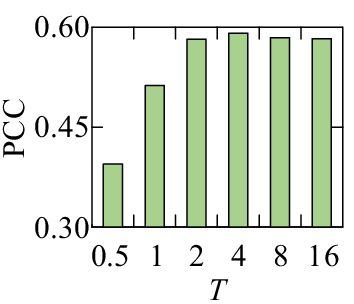}
}\hspace{-3.5mm}
\subfigure[Case 3]{
 \includegraphics[width=1.17in, height=1.12in]{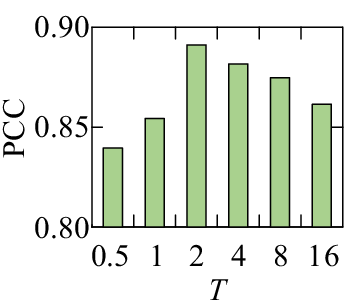}
}
\vspace{-2mm}
\caption{{\textsf{MMQ} vs. $T$}}
\vspace{-6mm}
\label{fig:E1.5}
\end{figure}

\vspace{-1mm}
\subsubsection{Impact of Parameter $T$}\label{ImpactofT}
As $T$ affects the soft labels, it may also affect the result of \textsf{MMQ}. In order to better study the influence of $T$ on \textsf{MMQ}, we compare the performance of \textsf{MMQ} over different $T$ values {for Cases 1-3}. 
{The top-1 model of case 1 is removed for a better view of the effect.}
Fig.~\ref{fig:E1.5} shows the PCC of \textsf{MMQ} by varying $T$ from 0.5 to 32.
It is observed that PCC achieves the highest value when $T$ is set around 2.0 for all three cases. Hence, it is better to set $T$ to 2.0 in practice. Also, the PCC values are high over most of the $T$ values if $T$ is neither too small (e.g., 0.5) nor too big (e.g., 32), which confirms the robustness of \textsf{MMQ} over different $T$ values.

\begin{figure}[t]
\centering
\includegraphics[width=0.46\textwidth]{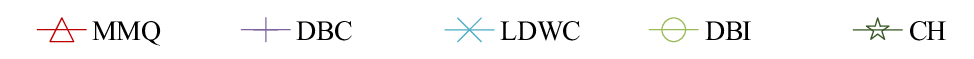}

\centering
\vspace{-2mm}
\hspace{-5mm}
\subfigure[Case 1]{
    \includegraphics[width=1.17in, height=1.12in]{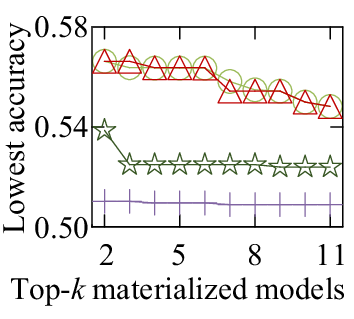}
}\hspace{-3.5mm}
\subfigure[Case 2]{
 \includegraphics[width=1.17in, height=1.12in]{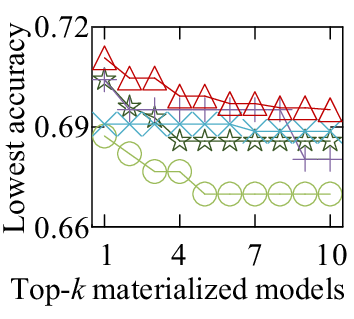}
}\hspace{-3.5mm}
\subfigure[Case 3]{
 \includegraphics[width=1.17in, height=1.12in]{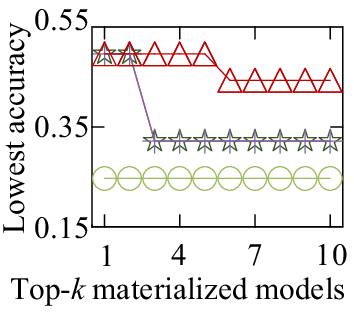}
}

\vspace{-2mm}
\caption{{\textsf{MMQ} vs. Clustering Evaluation Metrics}}
\vspace{-6mm}
\label{fig:E2.8}
\end{figure}

\vspace{-1mm}
\subsubsection{Comparison with Clustering Evaluation Metrics}
We compare our separation degree metric with four popular clustering evaluation metrics, including the distance between centroids (DBC for short), the longest distance within cluster (LDWC for short), Davies–Bouldin index~\cite{davies1979cluster} (DBI for short), and Calinski-Harabasz score~\cite{calinski1974dendrite} (CH for short), using Cases 1-3 stated in Section~\ref{sub:setup}.

Fig.~\ref{fig:E2.8} shows the lowest accuracy of top-$k$ materialized models ranked by \textsf{MMQ} using the separation degree and four different clustering evaluation metrics.
Note that, LDWC metric calculation is time-consuming, especially when the target dataset is large. We omit the results of LDWC for Cases 1 and 3 in Fig.~\ref{fig:E2.8}(a) and \ref{fig:E2.8}(c), as it can not finish computing within 24 hours.
It is observed that the separation degree always performs the best, which validates the effectiveness of the separation degree.
\begin{figure}[t]
\centering
\subfigure[\textsf{MMQ}]{
\includegraphics[width=0.21\textwidth]{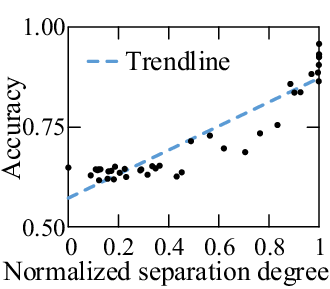}
}
\hspace{1.4mm}
\hspace{-0.25cm}
\subfigure[KLD]{
\includegraphics[width=0.21\textwidth]{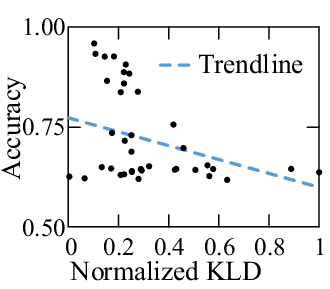}
}
\vspace{-1mm}
	
\centering
\hspace{-0.25cm}
\subfigure[JSD]{
\includegraphics[width=0.21\textwidth]{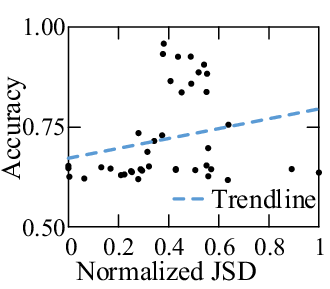}
}
\hspace{1.4mm}
\hspace{-0.25cm}
\subfigure[ALL]{
\includegraphics[width=0.21\textwidth]{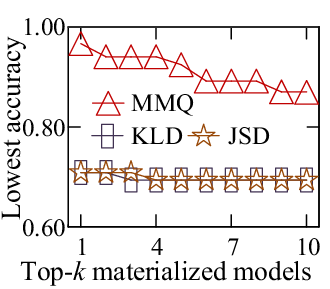}
}
\vspace*{-2mm}
\caption{{\textsf{MMQ} vs. KLD and JSD}}
\vspace*{-6mm}
\label{fig:E1.1}
\end{figure}

\subsubsection{Comparison with the Method Using Meta-data}\label{sub:metadata}
In this set of experiments, we explore the effect of data distribution (i.e., one type of meta-data) to find the best (i.e., the closest) materialized model from the target data.
We use Kullback–Leibler divergence (KLD)~\cite{kld} and Jensen–Shannon divergence (JSD)~\cite{jsd} to measure the similarity between the data distribution of the source dataset and that of the target dataset.
Next, we compare our method \textsf{MMQ} with KLD and JSD using CIFAR-10 dataset.

Specifically, we randomly divide CIFAR-10 dataset into two sub-datasets, where one sub-dataset that contains 10\% data is used as the target dataset, while the other sub-dataset contains the rest 90\% data. Then, we generate 40 materialized models with VGG-11 architecture from the sub-dataset with 90\% data, and randomly select 10-20\% data for training.
To get the ground truth accuracy, we use the same experimental setup as Case 1.
Fig.~\ref{fig:E1.1}(a) to \ref{fig:E1.1}(c) plot the accuracy of each materialized model retrained on the target dataset w.r.t. different model query metrics.
As observed, our method \textsf{MMQ} has the strongest correlation between our model query metric and the ground truth accuracy, i.e., models with higher normalized separation degree achieve higher accuracy.
Fig.~\ref{fig:E1.1}(d) shows that the top-$k$ query results of \textsf{MMQ} are stably better than those of KLD and JSD with the growth of $k$.
Hence, \textsf{MMQ} can rank materialized models better compared with KLD or JSD.

The method using meta-data does not perform well because the meta-data is utilized to measure the similarity between source dataset and target dataset.
However, the similarity between source dataset and target dataset cannot determine the result of model reuse completely, as proved by previous studies.
Except for the similarity, there are many other factors that determine how much useful knowledge the materialized model has for the target task, such as the way of training (e.g., epoch, learning rate, etc.). Thus, \textsf{MMQ}  directly measures the target-related knowledge of materialized models, and performs materialized model query better.

\begin{table}[]
\centering
\caption{{Average Running Time Except Predicting in Case 1 and Case 5}}
\vspace{-3mm}
\label{tab:E4-5}
\renewcommand\arraystretch{1.1}
\begin{tabular}{c|c|c|c|c|c|c}
\hline
Data                   & Dim & \multicolumn{1}{c|}{I-MMQ} & \multicolumn{1}{c|}{\begin{tabular}[c]{@{}c@{}}LLE\\ (n=5)\end{tabular}} & \multicolumn{1}{c|}{\begin{tabular}[c]{@{}c@{}}LLE\\ (n=20)\end{tabular}} & \multicolumn{1}{c|}{\begin{tabular}[c]{@{}c@{}}Isomap\\ (n=5)\end{tabular}} & \multicolumn{1}{c}{\begin{tabular}[c]{@{}c@{}}Isomap\\ (n=20)\end{tabular}} \\ \hline
\multirow{2}{*}{Case 1} & 5  & 0.127                      & 122.400                                                                  & 173.763                                                                   & 172.272                                                                     & 248.121                                                                      \\ \cline{2-7} 
                       & 25  & 0.148                      & 395.834                                                                  & 395.841                                                                   & 555.954                                                                     & 629.098                                                                      \\ \hline
\multirow{2}{*}{Case 5} & 5  & 0.464                      & 0.824                                                                    & 1.141                                                                     & 2.368                                                                       & 3.336                                                                        \\ \cline{2-7} 
                       & 25  & 0.666                      & 1.133                                                                    & 1.216                                                                     & 2.774                                                                       & 3.710                                                                        \\ \hline
\end{tabular}
\end{table}

\begin{figure}[t]
\vspace*{-4mm}
\centering
\includegraphics[width=0.46\textwidth]{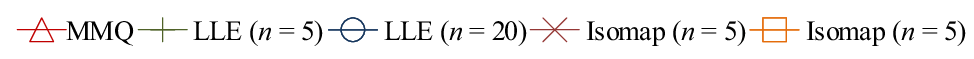}

\centering
\vspace{-2mm}
\hspace{-5mm}
\subfigure[\textsf{Isomap ($n=20$) in Case 1 with 5 output}]{
\includegraphics[width=0.21\textwidth]{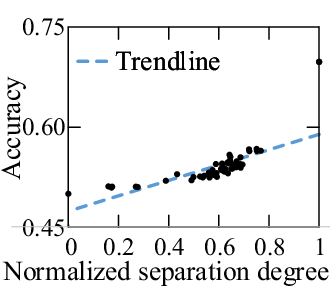}
}
\hspace{1.4mm}
\hspace{-0.25cm}
\subfigure[Isomap ($n=20$) in Case 5 with 5 output]{
\includegraphics[width=0.21\textwidth]{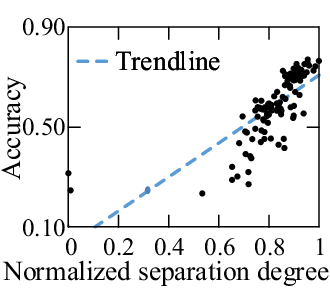}
}
\vspace{-1mm}

\centering
\vspace{-2mm}
\hspace{-5mm}
\subfigure[Case 1 with 5 output]{
\includegraphics[width=0.21\textwidth]{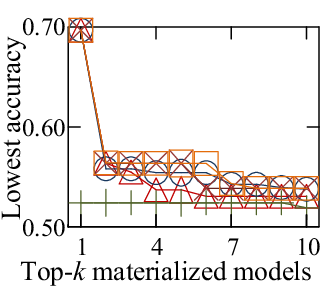}
}
\hspace{1.4mm}
\hspace{-0.25cm}
\subfigure[\textsf{Case 1 with 25 output}]{
\includegraphics[width=0.21\textwidth]{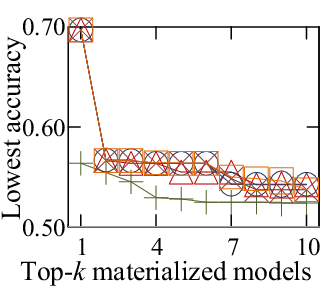}
}
\vspace{-1mm}
	
\centering
\hspace{-0.25cm}
\subfigure[Case 5 with 5 output]{
\includegraphics[width=0.21\textwidth]{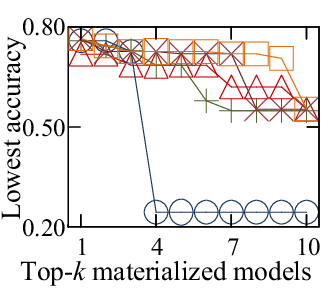}
}
\hspace{1.4mm}
\hspace{-0.25cm}
\subfigure[Case 5 with 25 output]{
\includegraphics[width=0.21\textwidth]{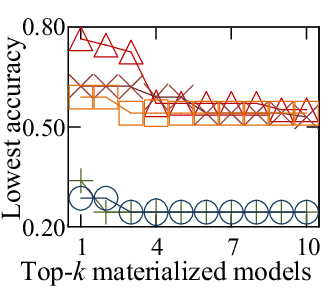}
}
\vspace*{-2mm}
\caption{{I-MMQ vs. LLE and Isomap}}
\vspace{-6mm}
\label{fig:nE1}
\end{figure}

\subsubsection{Comparison with Manifold Learning Methods}
\label{sec:Reduction}

{
We compare our I-MMQ by an additional fully connected layer with two popular manifold learning methods, i.e., LLE and Isomap, which are able to maintain distance relationships after reduction. We set the neighbors ($n$ in Fig.~\ref{fig:nE1}) of LLE and Isomap as 5 and 20. Since LLE and Isomap are time-consuming or occupy lots of memory, we sample 10\% of data quantity and 10\% of classes to calculate the separation degree. We set the dimensionalities after reduction 5 and 25. Fig.~\ref{fig:nE1} reports the correponding results. }

{
As shown in Fig.~\ref{fig:nE1}(a) and~\ref{fig:nE1}(b), Isomap ($n=20$) is able to rank materialized models even after reduction to 5 dimensions. Furthermore, Figs.~\ref{fig:nE1}(c)-\ref{fig:nE1}(f) show that Isomap can achieve good results in most of the time, while LLE can only achieve good results sometimes. Meanwhile, I-MMQ using fully connected layers always achieves comparable results. In addition, as shown in Table~\ref{tab:E4-5}, I-MMQ takes much shorter running time than LLE and Isomap methods. Therefore, it is reasonable to choose I-MMQ using fully connected layers in order to achieve high efficiency and comparable effectiveness.}

%% file: relatedwork.tex
\vspace{-2mm}
\section{Related Work}\label{sec:relatedwork}
\noindent\textbf{ML Model Management and Materialized Model Query.}
Recently, managing ML models has attracted lots of research interests. Several systems~\cite{DBLP:conf/icde/MiaoLDD17, DBLP:conf/icde/MiaoLDD17a, DBLP:conf/sigmod/VartakSLVHMZ16} are proposed to manage models during machine learning lifecycle. Besides, there exist studies about managing models for model diagnosis~\cite{DBLP:conf/sigmod/VartakTMZ18} and model governance in production ML~\cite{DBLP:conf/usenix/SridharSASRT18}. Recent system~\cite{DBLP:conf/sigmod/DerakhshanMARM20} also focuses on collaborative environment, and presents a system to optimize the execution of ML workloads in collaborative environments by reusing previously performed operations and their results while \cite{DBLP:conf/sigmod/GharibiWARL19} seek to automate the management of deep learning experiments.
Some ML model management systems~\cite{DBLP:conf/icde/MiaoLDD17, DBLP:conf/sigmod/00010DD17} provide the function of reusing materialized models by transfer learning. Those systems also offer materialized model query methods to enable users to find the most appropriate materialized models from model repositories. However, these methods rely on querying keywords and datasets that are uploaded by users to find materialized models trained on semantic-relevant tasks, which makes them lack of privacy protection. Also, the methods are ineffective since plenty of previous studies have shown that the semantic relevance between the materialized model and the target task does not determine the result of transfer learning~\cite{DBLP:journals/pr/AfridiRS18, DBLP:conf/nips/NeyshaburSZ20}.

Besides ML model management systems with materialized model query methods, a lot of materialized model query methods have been proposed in recent years.
\cite{DBLP:conf/ijcai/XiangPPSY11} utilizes context information from WWW or Wikipedia to build a bridge between the source and the target domain to complete the transfer of knowledge. However, this method depends on source data, and cannot support the situation when the source labels and the target labels are completely different.
Recently, \cite{DBLP:journals/pr/AfridiRS18} presents a materialized model query framework based on information theory. Nonetheless, this framework is time-consuming due to the training process. In addition, the framework can only be applied to specific types of models, because it relies on the operation of the network structure.
More recently, \cite{DBLP:conf/icde/Sigl19} provides a method for materialized model query based on the similarity between source dataset and target dataset, which avoids model training. 
Nonetheless, it needs the uploading of the source training datasets, which is time-consuming and limits its applicability.


\vspace{0.1in}
\noindent\textbf{Data Management Challenges in ML.}
Nowadays, a lot of efforts~\cite{DBLP:conf/sigmod/MiaoNSYJM021, DBLP:journals/vldb/GuoZJWZCL21, DBLP:journals/pvldb/LiSZJLDZY00021, DBLP:journals/pvldb/MiaoZSNYTC21, DBLP:journals/pvldb/AdnanMMN21} have been devoted to tackling data management-related challenges in machine learning.
Those studies leverage the technologies and concepts that are widely used in the data management and database research areas.
Some of them~\cite{DBLP:journals/pvldb/ElgoharyBHRR16, DBLP:conf/sigmod/LiCZ00NP19} use database compression techniques to speed up machine learning operations.
Similarly, SketchML~\cite{DBLP:conf/sigmod/JiangFY018} accelerates distributed machine learning based on the concept of sketch.
Recent work~\cite{DBLP:conf/sigmod/NakandalaKP19} also utilizes the concepts of incremental view maintenance and multi-query optimization to boost CNN predictions in occlusion based explanations.
Besides, the concepts of data provenance and cleaning are used to incrementally update model parameters~\cite{DBLP:conf/sigmod/WuTD20}.
This paper also tries to solve a data management problem in machine learning, i.e., materialized model query problem, and we construct a novel metric to find appropriate materialized models.
More recently, \cite{DBLP:journals/pvldb/ZhangYWSLW021} connects data selection with social influence maximization and improves the efficiency and performance of data selection in GNNs.

Transfer learning, which aims to improve the learning performance with one given materialized model, is also related to our work. Our work aims at finding the most appropriate materialized model. Then, the transfer learning methods do their jobs and try their best to leverage our selected materialized model. Thus, materialized model query problem is orthogonal to transfer learning. 

%% file: conclusion.tex
\vspace{-3mm}
\section{Conclusion}\label{sec:conclusions}
In this paper, considering the limitations of existing studies, we present a source-data free, general, efficient, and effective materialized model query framework \textsf{MMQ}. \textsf{MMQ} can efficiently rank materialized models for different tasks under different model structures without training and any source data, and is also able to work with various model structures simultaneously.
Given a materialized model, \textsf{MMQ} first utilizes soft labels to represent the samples in the target dataset, and uses Gaussian distributions to fit clusters of soft labels. Then, \textsf{MMQ} utilizes the separation degree on the Gaussian distributions to measure the target-related knowledge of the materialized model.
In addition, an improved \textsf{MMQ} is proposed to further reduce the query time while retaining the query performance. Finally, we conduct extensive experiments to demonstrate the effectiveness, efficiency, and generality of \textsf{MMQ}.


%% file: acknowledgments.tex
\vspace{-3mm}
\section*{Acknowledgments}

The authors would like to acknowledge Yuren Mao from Zhejiang University for his wonderful collaboration and patient support. Also, this work was supported by the NSFC under Grants No. (62025206, 61972338, and 62102351). Lu Chen is the corresponding author of the work.